\documentclass[11pt]{article}

\usepackage[final]{acl}

\usepackage{times}
\usepackage{latexsym}
\usepackage{float}
\usepackage{pifont}
\newcommand{\cmark}{\ding{51}} 
\newcommand{\xmark}{\ding{55}} 

\usepackage[T1]{fontenc}

\usepackage[utf8]{inputenc}

\usepackage{microtype}

\usepackage{inconsolata}

\usepackage{graphicx}
\usepackage{inconsolata}
\usepackage{booktabs}
\usepackage{soul}
\usepackage{color}

\usepackage{multirow}
\usepackage{multicol}
\usepackage{enumitem,kantlipsum}
\usepackage[normalem]{ulem}

\usepackage{color,colortbl}
\usepackage{todonotes} 
\usepackage{soul}
\usepackage{threeparttable, makecell}
\usepackage{makecell}
\usepackage{xcolor}

\usepackage{comment}
\usepackage{array}  
\usepackage{lscape}  
\usepackage{listings} 
\usepackage{multirow}
\usepackage{graphicx}
\usepackage[normalem]{ulem}
\useunder{\uline}{\ul}{}
\usepackage{amsmath}
\usepackage{array}  
\usepackage{multirow}  
\usepackage{amsmath}
\usepackage{subcaption}
\usepackage{fancybox}
\usepackage{tikz}

\usepackage{arabtex}
\usepackage{utf8}
\setcode{utf8}

\definecolor{codegray}{gray}{0.9}

\lstdefinestyle{pythonstyle}{
    backgroundcolor=\color{codegray},   
    language=Python,
    basicstyle=\ttfamily\footnotesize,
    keywordstyle=\color{blue},
    stringstyle=\color{red},    
    breaklines=true,
    frame=single,
    keepspaces=true,
    showstringspaces=false,
}

\lstdefinestyle{aclprompt}{
  basicstyle=\ttfamily\small,
  columns=fullflexible,
  breaklines=true,
  breakatwhitespace=true,
  showstringspaces=false,
  keepspaces=true,
  frame=single,
  framerule=0.4pt,
  rulecolor=\color{black!25}
}

\lstdefinestyle{promptstyle}{
  basicstyle=\ttfamily\footnotesize,
  breaklines=true,
  breakatwhitespace=false,
  columns=fullflexible,
  keepspaces=true,
  showstringspaces=false,
  frame=single,
  framerule=0.3pt,
  rulecolor=\color{black!25},
  xleftmargin=0.5em,
  xrightmargin=0.5em,
  aboveskip=0.6em,
  belowskip=0.6em
}

\usepackage{graphicx}
\definecolor{lightblue}{rgb}{.50,.95,1}
\definecolor{tri}{rgb}{.25,.88,.82}
\definecolor{lilac}{rgb}{0.85,0.64,0.85}

\newcommand{\sadiq}{\emph{Fanar-Sadiq}}

\usepackage{ulem}
\usepackage{hyperref}
\usepackage{verbatim}
\usepackage{setspace}

\usepackage{colortbl}
\usepackage{booktabs}

\usepackage[edges]{forest}
\usepackage{forest}

\usetikzlibrary{arrows.meta}
\tikzset{>=Latex}

\usepackage{xcolor}
\definecolor{dkgreen}{rgb}{0,0.5,0}
\definecolor{dkgreen}{rgb}{0,0.5,0}
\definecolor{gray}{rgb}{0.5,0.5,0.5}
\definecolor{mauve}{rgb}{0.58,0,0.82}

\lstdefinestyle{pythonstyle}{
  frame=tb,
  language=python,
  aboveskip=3mm,
  belowskip=3mm,
  showstringspaces=false,
  columns=flexible,
  basicstyle={\small\ttfamily},
  numbers=none,
  numberstyle=\tiny\color{gray},
  keywordstyle=\color{dkgreen},
  commentstyle=\color{dkgreen},
  breaklines=true,
  breakatwhitespace=true,
  tabsize=3,
  escapeinside={`}{`},
  breakindent=0pt,
  otherkeywords={these}
}

\usepackage{array}  
\usepackage{multirow}  
\usepackage{geometry}  
\usepackage{amsmath}

\usepackage{algorithm}
\usepackage{algorithmicx}
\usepackage{algcompatible} 

\usepackage{subcaption}
\usepackage{fancybox}
\usepackage{tikz}

\definecolor{dkgreen}{rgb}{0.0,0.45,0.0}
\definecolor{lstgray}{rgb}{0.5,0.5,0.5}

\usepackage{tikz}
\usetikzlibrary{arrows.meta,positioning,shapes,calc}

\usepackage{amsmath} 

\lstdefinestyle{pythonstyle}{
  frame=tb,
  language=Python,
  aboveskip=3mm,
  belowskip=3mm,
  showstringspaces=false,
  columns=fullflexible,
  basicstyle=\small\ttfamily,
  numbers=none,
  numberstyle=\tiny\color{lstgray},
  keywordstyle=\color{dkgreen},
  commentstyle=\color{dkgreen},
  breaklines=true,
  breakatwhitespace=true,
  tabsize=3,
  breakindent=0pt,
  escapeinside={(*@}{@*)},
  otherkeywords={}
}

\newcommand{\AR}[1]{\begin{RLtext}#1\end{RLtext}}

\title{Multiagent System for Islamic Related QA}
\title{A Multi-Agent System for Grounded Islamic QA}
\title{A Multi-Agent Architecture for Grounded Islamic QA}
\title{Fanar-Sadiq: A Multi-Agent Architecture for Grounded Islamic QA}

\author{
Ummar Abbas,
Mourad Ouzzani, 
Mohamed Y. Eltabakh, \\
\textbf{Omar Sinan,
Gagan Bhatia,
Hamdy Mubarak,
Majd Hawasly,} \\
\textbf{Mohammed Qusay Hashim,}
\textbf{Kareem Darwish,}
\textbf{Firoj Alam} \\
Qatar Computing Research Institute, HBKU, Qatar \\
\texttt{\{uabbas, mouzzani, meltabakh, osinan, hmubarak, mhawasly\}@hbku.edu.qa}\\
\texttt{\{mohashim, kadarwish, fialam\}@hbku.edu.qa, gbhatia@qcri.org}
}

\begin{document}
\maketitle

\begin{abstract}
Large language models (LLMs) can answer religious knowledge queries fluently, yet they often hallucinate and misattribute sources, which is especially consequential in Islamic settings where users expect grounding in canonical texts (Qur'an and Hadith) and jurisprudential (fiqh) nuance. Retrieval-augmented generation (RAG) improves grounding, however, a single retrieve-then-generate pipeline is insufficient for diverse Islamic queries, including verbatim scripture, citation-grounded guidance, and rule-constrained computations such as zakat and inheritance. To address these challenges, we present \textit{Fanar-Sadiq}, a bilingual Arabic-English Islamic QA system built on a multi-agent, tool-augmented architecture.\footnote{It is a core component of the \href{https://fanar.qa/en}{Fanar AI platform}~\cite{fanar2024}.}
\textit{Fanar-Sadiq} routes Islamic queries to specialized modules within an agentic tool architecture. It supports intent-aware routing, retrieval-grounded fiqh answers with normalized citations and verification traces, exact verse lookup with quotation validation, and deterministic Sunni zakat and inheritance calculators with madhhab-sensitive branching. We evaluate the end-to-end system on public Islamic QA benchmarks and show strong effectiveness and efficiency. It is publicly accessible through an API and Web application and has received over 1.9M accesses in less than a year.\footnote{\url{https://api.fanar.qa/docs}}
\end{list} 
\end{abstract}

\section{Introduction}
Recent advances in large language models have enabled conversational assistants that can handle knowledge-intensive question answering (QA) across many domains~\cite{ma-etal-2025-large-language-models-meet,hasan-etal-2025-nativqa}. Despite these gains, hallucination and source-attribution errors remain common, particularly when users expect answers grounded in authoritative references rather than plausible-sounding narrative~\cite{chen2026survey}.
In religious applications, such failures are high-stakes. Fabricated Quranic verses, misattributed Hadith, or unqualified jurisprudential claims can mislead users~\cite{Bhatia_2026}. This motivates Islamic QA systems that provide grounded answers, stable citations, and explicit handling of abstention and scholarly disagreement.


\begin{figure*}[!t]
    \centering
    \includegraphics[width=0.95\linewidth]{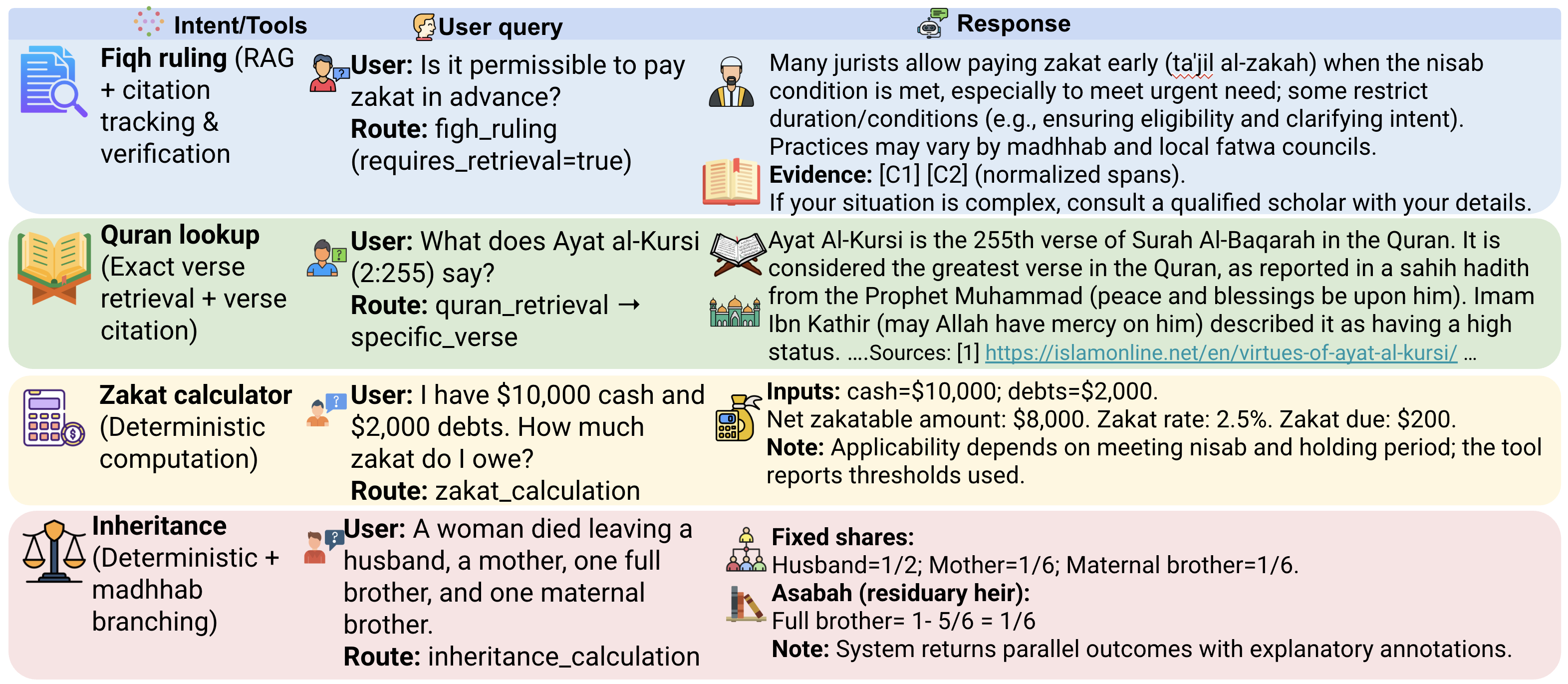}
    \vspace{-0.3cm}
    \caption{Illustrative end-to-end examples showing intent routing to specialized tools, traceable citations for fiqh QA, exact Quranic verse handling, deterministic zakat computation, and explicit madhhab-sensitive branching for disputed inheritance cases. Citation tags [C*] denote normalized evidence spans: [Q*] denotes verse-level citations.}
    \label{fig:fanar_sadiq_pipeline_example}
    \vspace{-0.4cm}
\end{figure*}

%
The community has begun formalizing these reliability requirements through benchmarks and shared tasks.  QuranQA~\citep{malhas2023quranqa} has established standardized evaluation for Quranic passage retrieval and reading comprehension. 
IslamicEval~\citep{mubarak-etal-2025-islamiceval} further emphasizes grounded Quran/Hadith QA and includes tasks for detecting and correcting Quranic hallucinations.
For structured religious reasoning, QIAS focuses on Islamic inheritance (aka faraid in Islamic Jurisprudence), a domain where correct answers require rule-based computation and legal constraints \citep{bouchekif2025qias}.

Islamic QA systems should 
\textit{(i)}~ground responses in canonical sources, 
\textit{(ii)}~provide transparent citations, and 
\textit{(iii)}~properly handle domain-specific reasoning tasks that exceed the reliability of free-form generation~\cite{Bhatia_2026}.
RAG combines parametric generation with non-parametric retrieval to ground responses in external 
documents~\citep{bhatia2026ragagenticragfaithful,lewis2020rag}; 
dense retrievers such as DPR~\citep{karpukhin2020dpr} and generative readers such as Fusion-in-Decoder (FiD)~\citep{izacard2021fid} (as well as retrieval-in-the-loop models such as RETRO~\citep{borgeaud2022retro} and Atlas \citep{izacard2023atlas}) show retrieval can improve factual QA and knowledge updating.
Nevertheless, a fixed \textit{retrieve-then-generate} pipeline is often not a good match for real-world Islamic queries; some are best satisfied by \emph{exact lookup}, e.g., ``What does verse 2:255 say?'', others require \emph{rule-constrained computation}, e.g., zakat, inheritance, and others require \emph{jurisprudential reasoning with evidence presentation}, e.g., fatwa-style questions with stated assumptions, conditions, and madhhab sensitivity. Thus, treating heterogeneous intents uniformly can degrade correctness and user experience. Tool-using approaches suggest a path beyond rigid pipelines. ReAct~\citep{yao2023react} interleaves reasoning with actions for iterative retrieval, while Toolformer~\citep{schick2023toolformer} shows that models can learn when to call tools e.g., calculators. These approaches motivate an architecture that selects execution modes based on query intent.

\begin{figure*}[!tbh]
    \centering
    \includegraphics[width=0.98\linewidth]{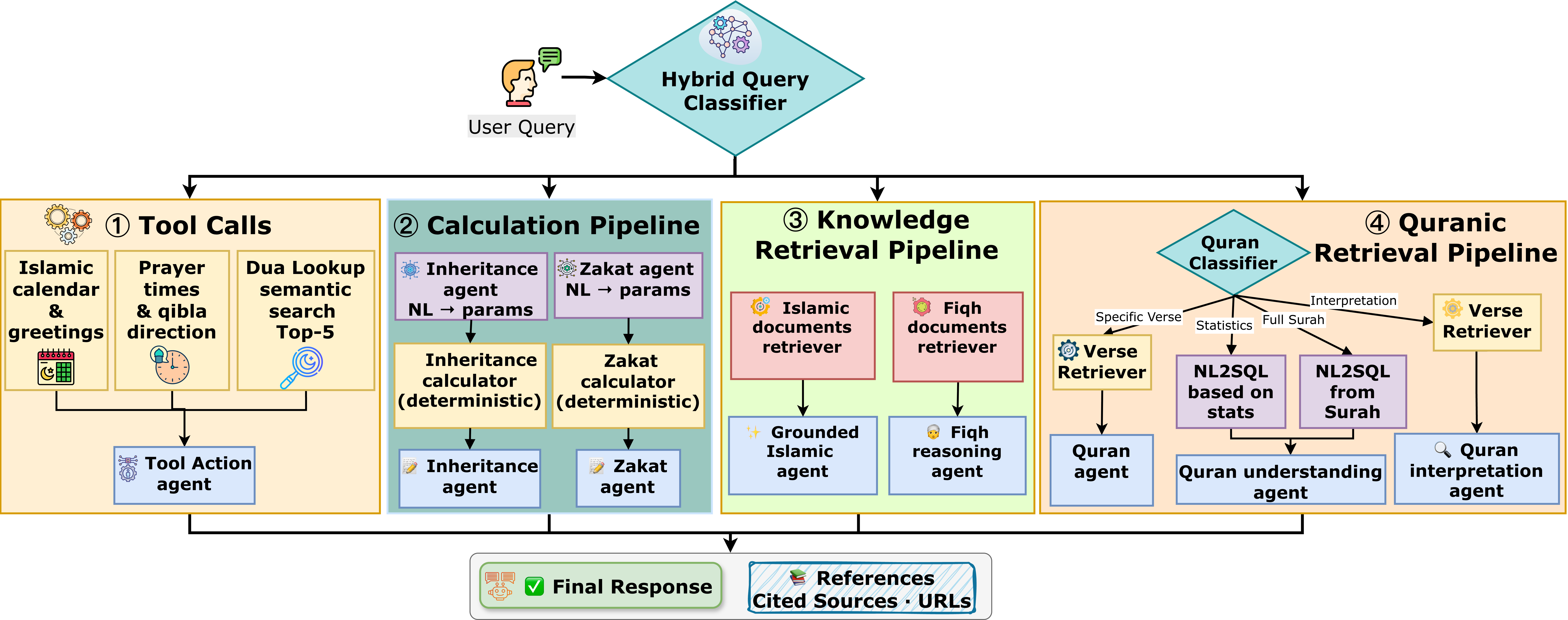}
    \vspace{-0.2cm}
    \caption{Our \textbf{\textit{multi-agent architecture}}. A hybrid query classifier selects among \textit{(i)} tool calls, \textit{(ii)} deterministic calculation, \textit{(iii)} document-grounded retrieval QA, and \textit{(iv)} Quranic retrieval routes, before assembling the final response with references.}
    \label{fig:fanar_sadiq_system}
    \vspace{-0.3cm}
\end{figure*}

Guided by this principle, in this paper, we present \textbf{\textit{Fanar-Sadiq}} a bilingual \textbf{\textit{multi-agent Islamic QA system}} built around an agentic, multi-tool architecture (see Figure~\ref{fig:fanar_sadiq_system}). It operationalizes intent-aware execution by first classifying Islamic queries into fine-grained intent types, including exact scripture lookup, retrieval-grounded jurisprudential QA, and rule-constrained computation. Examples are shown in Figure~\ref{fig:fanar_sadiq_pipeline_example}.
It then routes each query to the appropriate specialized module and enforces transparency and reliability through citation tracking and post-generation verification.

This design enables intent-aware, tool-routed execution as a reliable alternative to fixed RAG for Islamic QA. Our contributions are:
\begin{itemize}[noitemsep,topsep=0em,leftmargin=1.5em,labelsep=.5em]
  \item A \textit{multi-agent architecture} for Islamic QA that goes beyond fixed RAG by routing queries to specialized tools and integrating evidence tracking and verification.
  \item A comprehensive evaluation spanning multiple public benchmarks covering both generative and multiple-choice Islamic QA.
  \item Our findings show that tool- and evidence-routed execution improves faithfulness, vital for Islamic QA, while remaining competitive on broader Islamic knowledge benchmarks.
\end{itemize}
\section{Related Work}

\noindent\textbf{Multi-Agent Tools for QA Systems.}
While RAG has established itself as the de-facto standard for knowledge-intensive NLP, mitigating hallucination via dense retrieval mechanisms \cite{borgeaud2022retro}, 
standard \textit{retrieve-then-generate} pipelines often struggle with heterogeneous user intents that demand multi-step reasoning or precise computation rather than mere semantic similarity \cite{fanar2024, bragg2025astabenchrigorousbenchmarkingai}. To address these structural limitations, the field has pivoted toward ``Agentic RAG'' and tool-augmented  models \cite{comanici2025gemini, bhatia2026ragagenticragfaithful}, where frameworks like ReAct \cite{yao2023react} enable models to interleave reasoning traces with external API calls to iteratively refine answers.
This paradigm shift is particularly critical for domain-specific applications. Recent work demonstrates that decomposing complex queries, such as mathematical reasoning \cite{shao2024deepseekmathpushinglimitsmathematical} or legal judgment \cite{bahaj2025mizanqa}, into modular subtasks handled by specialised agents significantly outperforms monolithic generation. 

\noindent\textbf{Islamic RAG Systems.}
The deployment of LLMs in the Islamic domain is constrained by the critical necessity of doctrinal integrity, where hallucination risks \cite{alansari2025arahallueval} and ``sacred versus synthetic'' attribution failures \cite{atif2025sacred} differ fundamentally from open-domain issues. 
Consequently, recent shared tasks such as QuranQA \cite{malhas2023quranqa} and IslamicEval 2025 \cite{mubarak-etal-2025-islamiceval} have established benchmarks for Islamic QA tasks where different LLMs based approaches has been explored~\cite{tajrin-etal-2025-aya}.
Initiatives like QIAS 2025 \cite{bouchekif2025qias} and Hajj-FQA \cite{aleid2025hajjfqa} explicitly target structured reasoning in inheritance and ritual jurisprudence. Beyond static benchmarks, architectural innovations are increasingly integrating reliability controls. Systems like AFTINA \cite{mohammed2025aftina} and FARSIQA \cite{asl2025farsiqa} employ RAG-based reranking and iterative refinement to ground Fatwa answers, while others leverage morphological constraints \cite{AHJ25} and cross-lingual augmentation \cite{oshallah2025crosslanguage} to ensure recitation accuracy. Most relevant to our approach, \cite{bhatia2026ragagenticragfaithful} introduced an agentic framework that utilizes structured tool calls for verse-level verification, demonstrating that iterative evidence seeking significantly reduces hallucination compared to standard RAG. 

\section{System Architecture}
\label{sec:system_design}


In Figure~\ref{fig:fanar_sadiq_system}, we present our \textit{multi-agent end-to-end architecture}. The system is designed for heterogeneous Islamic QA, spanning rule-heavy obligations best handled with symbolic computation and canonical text retrieval where verbatim accuracy is essential. User queries fall into three broad classes: 
\textit{(i)}~text-grounded questions (Quran/Hadith/fiqh/general Islamic knowledge), 
\textit{(ii)}~rule- and arithmetic-constrained questions (zakat and inheritance), and 
\textit{(iii)}~symbolic time/geo questions (Hijri calendar and prayer times). Treating all of these intents as a single \textit{retrieve-then-generate} task leads to predictable failure modes, including misquoted verses, weak or missing sourcing for jurisprudential claims, and numerically inconsistent zakat or inheritance outputs.
To contextualize these design choices, in Table~\ref{tab:arch_comparison}, we compare our system with prior agentic and Islamic QA systems, highlighting why Islamic QA benefits from specialized modules for computation and scripture handling.



Most Islamic QA systems use text-only retrieval-generation workflows, sometimes with reranking, iterative retrieval, or transparent citations, as in AFTINA, FARSIQA, and MufassirQAS~\cite{mohammed2025aftina,alan2025improving}. Yet they largely treat heterogeneous Islamic queries as a single \textit{retrieve-then-generate} task. Agentic RAG improves faithfulness through structured evidence seeking and answer revision, however, mainly extends retrieval behavior rather than unifying deterministic jurisprudential calculators and symbolic geo-temporal tools under intent-aware routing~\cite{bhatia2026ragagenticragfaithful}.

In contrast, \textbf{our multi-agent architecture} routes each query to a heterogeneous tool suite, including deterministic zakat and inheritance calculators (Figure~\ref{fig:fanar_sadiq_inheritance_system} in Appendix), canonical verse lookup, and rule-based calendar and prayer-time computation. We further apply citation normalization and post-generation verification to reduce Quran/Hadith misquotation and support rule-intensive inheritance reasoning. In our multi-agent system, we use Fanar~\cite{fanar2024} as the LLM agent. We chose it because it offers free API access 
and has demonstrated strong performance across diverse benchmarks~\cite{bhatia2026ragagenticragfaithful}.

\begin{table}[t]
\centering
\small
\setlength{\tabcolsep}{2pt} 
\scalebox{0.75}{%
\begin{tabular}{l c c c c c}
\hline
\textbf{Work} &
\textbf{Calc.} &
\textbf{Quran} &
\textbf{NL2SQL} &
\textbf{Evidence} &
\textbf{Tools} \\
\hline
\rowcolor{teal!30}
\textbf{Ours (Fig.~\ref{fig:fanar_sadiq_system})} &
\cmark & \cmark & \cmark & \cmark & \cmark \\
\hline
\citet{karpas2022mrkl} (MRKL) &
\xmark & \xmark & \xmark & \cmark & \cmark \\
\citet{yao2023react} (ReAct) &
\xmark & \xmark & \xmark & \cmark & \cmark \\
\citet{schick2023toolformer} (Toolformer) &
\xmark & \xmark & \xmark & \cmark & \cmark \\
\citet{asai2024selfrag} (Self-RAG) &
\xmark & \xmark & \xmark & \cmark & \xmark \\
\citet{yan2024crag} (CRAG) &
\xmark & \xmark & \xmark & \cmark & \xmark \\
\citet{alazani2025ontologyragq} &
\xmark & \cmark & \xmark & \cmark & \xmark \\
\citet{bhatia2026ragagenticragfaithful} &
\xmark & \cmark & \xmark & \cmark & \cmark \\
\citet{omayrah2025humain} &
\xmark & \cmark & \xmark & \cmark & \xmark \\
\citet{al-smadi-2025-qu} &
\xmark & \xmark & \xmark & \cmark & \xmark \\
\citet{alowaidi-2025-sea} &
\xmark & \xmark & \xmark & \cmark & \xmark \\
\bottomrule
\end{tabular}}
\vspace{-0.3cm}
\caption{Comparison against widely-recognized tool/agent/RAG architectures.
\textbf{Calc.} refers to explicit rule engines. \textbf{Quran} denotes verse-anchored retrieval distinct from generic document search. 
}
\label{tab:arch_comparison}
\vspace{-0.35cm}
\end{table}

\subsection{Hybrid Query Classifier} 
As shown in Figure~\ref{fig:fanar_sadiq_system}, we implement a hybrid routing classifier to predict the query type and select the execution route as the system entry point. The primary classifier is an LLM prompted to output an intent label, a confidence score, a short rationale, optional decomposition subquestions, and a retrieval flag indicating whether evidence retrieval is required. We define nine intent classes aligned with the system tools: \textit{(i)} fiqh rulings, \textit{(ii)} Qur'an retrieval, \textit{(iii)} general Islamic knowledge, \textit{(iv)} greetings/chitchat, \textit{(v)} zakat calculation, \textit{(vi)} inheritance calculation, \textit{(vii)} du'a (or supplication) lookup, \textit{(viii)} Islamic calendar, and \textit{(ix)} prayer times (see more details in Appendix (App.)~\ref{app:classifier}). 

Our nine intent classes are motivated by established query-intent and dialogue-act views that separate \textit{(a)} social acts (e.g., greetings), \textit{(b)} information-seeking retrieval (e.g., Quran text and dua lookup), and \textit{(c)} transactional/computational requests (e.g., zakat, inheritance, calendar, and prayer-time queries), where each intent implies a different execution strategy and error profile \citep{broder2002taxonomy}. We further ground the schema in canonical subdomains of Islamic knowledge: jurisprudential queries require fiqh-oriented reasoning \citep{hallaq2009intro}, while zakat and inheritance follow structured rule systems amenable to calculator-style tools \citep{qaradawi2000fiqhalzakah}. Calendar conversion and prayer-time/Qibla requests are naturally modeled as spatiotemporal computations \citep{reingold2018calendrical}, yielding a taxonomy that is both operationally tool-aligned and semantically coherent. We developed a manually annotated dataset of 700 queries to evaluate the hybrid classifier, which achieves 90.1\% accuracy. More details on the dataset development process and comparative results are provided in App.~\ref{ssec_app_query_classifier}.

\subsection{Tool Calls} 
Queries requiring utility lookups are routed to the \textit{Tool Action Agent}, which orchestrates deterministic modules. A lightweight \textbf{Greeting Tool} handles culturally appropriate greetings and pleasantries (App.~\ref{app:greeting_tool}). The \textbf{Islamic Calendar} tool performs rule-based time reasoning for Hijri date queries, Gregorian-Hijri conversion, and event lookups using multilingual intent cues, \texttt{hijridate} conversions, and a curated bilingual event ontology with explicit year-rollover logic, with a controlled fallback that warns about local moon-sighting variance (App.~\ref{app:islamic_calendar}). The \textbf{Prayer Times and Qibla} tool resolves locations to coordinates using a curated city database, with a rate-limited geocoding fallback. It computes prayer timetables using \texttt{pyIslam} with method-specific parameters. For Qibla requests, it computes the great-circle distance and bearing to Makkah, while logging trace metadata for interpretability (App.~\ref{app:prayer_times}). Finally, the \textbf{Dua Lookup} tool provides high-recall, deterministic retrieval. It first selects top-$k$ occasions using semantic search over precomputed embeddings. A lightweight LLM selector then maps the best match to canonical \texttt{page\_title} keys. The tool returns the supplication verbatim, including Arabic text, translation, and reference, from a structured store to avoid rewriting (App.~\ref{app:dua_lookup}).


\subsection{Calculation Pipeline (Deterministic)}
Rule-heavy financial and legal questions are routed to the \textit{Calculation Pipeline} to eliminate arithmetic drift and enforce jurisprudential constraints.

\noindent\textbf{Zakat Calculator.}
Zakat is a Shariah-mandated almsgiving governed by established juristic rules \citep{alqaradawi1999fiqh}. Our Zakat agent extracts structured parameters such as asset classes, amounts, and debts, then passes them to a deterministic module. The calculator computes the \emph{nisab} or minimum threshold based on precious-metal prices and applies category-specific logic.\footnote{\url{https://sunnah.com/bukhari:1483}} For agriculture, differentiated rates are applied based on irrigation methods. For livestock, Hadith-based schedules are used for camels, cattle, and sheep. For assets, rates are applied to cash, gold, business assets, and investments after deducting eligible debts. The output is a structured breakdown of inputs, deductions, and totals, formatted into a user-facing explanation with citations (App.~\ref{app:zakat_calculator}).

\noindent\textbf{Inheritance Calculator.}
The inheritance calculator (Figure~\ref{fig:fanar_sadiq_inheritance_system}) is a deterministic Sunni module that computes estate distribution while explicitly handling madhhab-specific differences. The workflow proceeds in three phases. First, fixed shares (\textit{fard}) are assigned to eligible heirs after validating kinship and removing impeded individuals. Second, the remaining estate is allocated via a priority chain of paternal-line relatives known as residuaries (\textit{`asaba}). Third, the module enforces arithmetic consistency by applying \textit{`awl} (proportional reduction) if shares exceed the estate, or \textit{radd} (return of remainder) if a surplus exists. Crucially, jurisprudentially disputed cases trigger a policy selector that returns parallel distributions, such as Hanafi versus Jumhur (majority opinion), rather than a single collapsed ruling (App.~\ref{app:inheritance_calculator}).

\subsection{Knowledge Retrieval Pipeline}
Informational queries are routed to retrieval-augmented QA workflows that instantiate \emph{usul al-fiqh} reasoning patterns through evidence linkage.

\noindent\textbf{Fiqh Rulings.}
This module uses a \textit{Fiqh Documents Retriever} followed by a reasoning agent. The agent is prompted to state ruling scope and assumptions, separate rulings from evidence, and assign deterministic citation tags to evidence spans. The system supports retrieval-time source normalization to ensure every claim maps to a stable source text. If a ruling relies on exact scriptural wording, the agent can invoke the Quranic tool to prevent paraphrase drift (App.~\ref{app:fiqh_tool}).

\noindent\textbf{General Islamic Understanding.}
For general inquiries, the system retrieves candidate documents and normalizes them into a bounded context. A \textit{Grounded Islamic Agent} then generates a response that is strictly grounded in the retrieved references to minimize hallucinations (App.~\ref{app:retriever}).

\subsection{Quranic Retrieval Pipeline}

\noindent\textbf{Quran Query Classifier.}
Quran-related queries are handled by a dedicated routing module that predicts one of four subtypes: \textit{specific verse}, \textit{full surah}, \textit{statistics}, or \textit{interpretation}. 
The primary classifier is an LLM constrained to this closed label set; if the output is invalid or non-conforming, the system falls back to an embedding-based classifier over exemplars to ensure stable routing under malformed outputs or low confidence.
The predicted subtype selects a fixed execution route and downstream response formatting, and the system logs structured metadata (predicted subtype, selected path, invoked tools) for traceability (App.~\ref{app:quran_tool}).

\noindent\textbf{Quran Interpretation and Specific Verse Retrieval.}
For \textit{specific verse} requests, the system invokes a Quran retrieval tool. These requests may include explicit \textit{surah:ayah} references, named surahs, or short quotations. The tool parses the reference and returns the canonical ayah text verbatim. If only partial information is provided, or parsing fails, it falls back to surah-level lookup.

For \textit{interpretation} queries, the system retrieves relevant verses and supporting documents. A constrained \textit{Quran Interpretation Agent} then produces an explanatory response grounded in the retrieved evidence. When references can be resolved, the system attaches verse-level citations to reduce paraphrase drift and ungrounded exegesis (App.~\ref{app:quran_tool}).

\noindent\textbf{NL2SQL for Complete Surah and Statistical Queries.} Requests requiring contiguous text or exact counting are routed to a NL2SQL module to avoid truncation, hallucination, and arithmetic errors.
%
For \textit{complete surah} queries, the system returns the complete chapter when feasible. Otherwise it executes \textit{SQL from Quran} to retrieve all verses in canonical order directly from the verse table. 
For \textit{statistics} queries, e.g., verse counts, word frequencies, surah metadata, and structural filters, the system executes \textit{SQL based on stats} to guarantee numerically exact outputs, optionally enriching numeric results with representative examples before formatting. Across both NL2SQL routes, the response renderer standardizes formatting, attaches citations/URLs, and records execution metadata for validation and debugging (App.~\ref{app:nl2sql_tool}).

All pipelines return a standardized output object with the natural language answer and structured metadata. A final \textit{response assembler} merges these results, adds a \textit{references block} with citations and URLs, and logs execution traces for validation and debugging (App.~\ref{app:response_assembly}).

\section{Evaluation}
\label{sec:methodology}

We evaluate our system end-to-end and compare it against strong proprietary and open-source baselines. Proprietary baselines include OpenAI models (GPT-4.1 and GPT-5) \cite{openai2023gpt4} and Google Gemini models (Gemini-3-Flash and Gemini-3-Pro) \cite{comanici2025gemini}. Open-source baselines include ALLaM-7B \cite{bari2024allamlargelanguagemodels} and Fanar-2-27B \cite{fanar2024}. Below, we briefly discuss benchmarking datasets. 

\subsection{Benchmarking Datasets}
\label{sec:eval_datasets}

We evaluate \sadiq{} on a suite of benchmarks covering two complementary settings as discussed below. 


\paragraph{Open-ended benchmarks.}
We use two open-ended benchmarks to assess generative reliability in settings where answers must be faithful, well-grounded, and contextually appropriate. 

\begin{itemize}[noitemsep,topsep=0em,leftmargin=1.5em,labelsep=.5em]
\item \textbf{IslamicFaithQA} contains \textit{3,810} bilingual Arabic-English examples with single-gold \emph{atomic} reference answers. It is designed to expose real-world failure modes in Islamic QA, including free-form hallucination and appropriate abstention when evidence is unavailable~\cite{bhatia2026ragagenticragfaithful}. 

\item \textbf{FatwaQA} is an Arabic benchmark of \textit{2,000} fatwa-style QA pairs focused on Islamic jurisprudence and finance, including \textit{zakat}, \textit{riba}, \textit{murabaha}, \textit{gharar}, \textit{waqf}, \textit{ijara}, \textit{maysir}, \textit{musharaka}, \textit{mudharaba}, \textit{takaful}, and \textit{sukuk}~\cite{fatwa_qa_evaluation}. Its open-ended format supports evaluation of detailed, evidence-backed responses under realistic prompts.
\end{itemize}

\paragraph{Multiple-choice benchmarks.}
We use three MCQ benchmarks to evaluate complementary aspects of Islamic QA, including factual religious knowledge, rule-constrained legal reasoning, and value-consistent decision making. These benchmarks allow controlled evaluation through exact-match scoring over predefined answer options.

\begin{itemize}[noitemsep,topsep=0em,leftmargin=1.5em,labelsep=.5em]

\item \textbf{QIAS 2025} focuses on Islamic inheritance reasoning, a hard-constraint fiqh task where models select the correct option corresponding to the gold inheritance distribution. We report exact-match accuracy over the selected option~\cite{bouchekif2025qias}. 

\item \textbf{PalmX 2025} Islamic Culture Subtask contains \textit{1,000} Arabic MSA multiple-choice questions covering Islamic culture and practices~\cite{palmx2025}. 

\item \textbf{IslamTrust} measures alignment with consensus-based Islamic ethical principles using a bilingual Arabic-English MCQ benchmark of \textit{406} items~\cite{lahmar2025islamtrust}.
\end{itemize}

In Table~\ref{tab:eval_datasets}, we outline the evaluation datasets, detailing their formats, supported languages, sizes, and corresponding evaluation metrics.

\noindent
\subsection{Evaluation method.}
For the open-ended datasets, IslamicFaithQA and FatwaQA, we use an \emph{LLM-as-a-judge} setup following \textsc{SimpleQA}~\cite{haas2025simpleqaverifiedreliablefactuality}. The judge LLM, GPT-4.1, receives the question, system response, reference answer, and evidence when available. It then assigns one of three verdicts, \emph{correct}, \emph{incorrect}, or \emph{not attempted}. The evaluation prompt is provided in App.~\ref{app:grader_template}. We aggregate these verdicts to report \%correct and abstention-aware reliability. For the MCQ datasets, PalmX, QIAS Subtask 1, and IslamTrust, we compute exact-match accuracy. The predicted option is compared against the gold label.

\begin{table}[t]
\centering
\small
\setlength{\tabcolsep}{3pt}
\renewcommand{\arraystretch}{1.12}
\scalebox{0.78}{
\begin{tabular}{l l l r l}
\toprule
\textbf{Dataset (Ref.)} & \textbf{Format} & \textbf{Lang} & \textbf{Size} & \textbf{Metric(s)} \\
\midrule
PalmX \citep{palmx2025}                          & MCQ   & ar     & 1,000 & Acc \\
QIAS (T1) \citep{bouchekif2025qias}       & MCQ   & ar     & 1,000 & Acc \\
IslamTrust \citep{lahmar2025islamtrust}          & MCQ   & ar+en  & 406  & Acc \\
IslamicFaithQA \citep{bhatia2026ragagenticragfaithful} & GenQA & ar+en & 3,810 & Acc (LLM-J) \\
FatwaQA \citep{fatwa_qa_evaluation}              & GenQA & ar     & 2,000 & Acc (LLM-J) \\
\bottomrule
\end{tabular}
}
\vspace{-0.2cm}
\caption{Evaluation datasets used in this work. Lang: \texttt{ar}=Arabic, \texttt{en}=English. GenQA: generative question answering. LLM-J: LLM-judge.}
\label{tab:eval_datasets}
\vspace{-0.2cm}
\end{table}

\section{Results \& Discussion}
\label{sec:results}

\begin{table}[t]
\centering
\setlength{\tabcolsep}{3pt}
\scalebox{0.65}{
\begin{tabular}{lccccccc}
\toprule
\textbf{Dataset} &
\textbf{GPT-4.1} &
\textbf{GPT-5} &
\textbf{G3-F} &
\textbf{G3-P} &
\textbf{ALLaM} &
\textbf{Fanar} &
\textbf{Ours} \\
\midrule
PalmX          & 52.9 & 82.3 & 81.2 & 84.4 & 45.5 & 72.5 & \textbf{85.5} \\
QIAS T1        & 89.2 & 93.0 & 91.5 & \textbf{94.5} & 52.4 & 63.5 & 72.2 \\
IslamTrust     & 94.7 & 95.2 & 94.8 & \textbf{95.6} & 57.4 & 83.2 & 94.2 \\
IslamicFaithQA & 41.4 & 51.2 & 53.4 & 56.6 & 42.7 & 48.2 & \textbf{65.4} \\
FatwaQA  & 32.3 & 63.6 & 54.6 & \textbf{67.0} & 31.5 & 44.5 & 65.1 \\
\midrule
Average        & 62.1 & 77.1 & 75.1 & \textbf{79.6} & 45.9 & 62.4 & 76.5 \\
\bottomrule
\end{tabular}
}
\vspace{-0.25cm}
\caption{Accuracy (\%) across benchmarks. G3-F: Gemini-3-Flash, G3-P: Gemini-3-Pro. 
}
\label{tab:main_results}
\vspace{-0.4cm}
\end{table}

In Table~\ref{tab:main_results}, we report accuracy across five benchmarks. \sadiq{} achieves an average accuracy of 76.5. It substantially outperforms the open-source baselines, ALLaM-7B at 45.9 and Fanar-2-27B at 62.4. It also remains competitive with strong proprietary models, including Gemini-3-Pro at 79.6 and GPT-5 at 77.1.

\sadiq{} shows the largest gains on \textbf{\textit{open-ended QA}}. On IslamicFaithQA, it reaches 65.4, compared with 56.6 for the strongest proprietary baseline. On FatwaQA, it achieves 65.1, closely matching Gemini-3-Pro at 67.0. These results support intent-aware routing. Instead of forcing all queries through a single \textit{retrieve-then-generate} pipeline, \sadiq{} selects specialized execution modes for different query types.

On \textbf{\textit{multiple-choice benchmark}}s, \sadiq{} performs strongly on broad Islamic knowledge, reaching 85.5 on PalmX. It also achieves 94.2 on IslamTrust, showing strong performance on value-sensitive decision making. These results suggest that the multi-tool design preserves general Islamic QA competence while improving reliability. 

QIAS Task 1 remains the most challenging benchmark. \sadiq{} obtains 72.2, while the strongest proprietary models reach 93.0-94.5. The MCQ format likely introduces an additional source of error. Even when \sadiq{} computes inheritance shares deterministically, it must still map the computed distribution to one of the benchmark's discrete answer options, which can lead to option-selection errors.

These results support our central hypothesis. Islamic QA benefits from intent-aligned execution rather than a uniform \textit{retrieve-then-generate} policy. Each module targets a different failure mode. Canonical verse lookup and quotation validation reduce paraphrase drift in scripture-related queries. Deterministic calculators enforce arithmetic and jurisprudential constraints for zakat and inheritance. Retrieval-grounded fiqh answering, combined with citation normalization, improves traceability and reduces unsupported claims.

Overall, these components explain the gains on open-ended benchmarks, where hallucination and attribution errors are strongly penalized. The results also reveal a limitation on QIAS-style MCQs. Although symbolic computation can produce correct inheritance shares, the system must still map them to constrained answer options. Future work should improve this symbolic-to-option alignment while preserving grounding and verification.

\section{Case Study: Chat Platform Integration}
We integrate \sadiq{} into the Fanar API platform as a specialized backend for a web-based chat interface. The platform uses an orchestrator to mediate incoming user queries. The orchestrator classifies each query and routes it to the appropriate component for execution. Concretely, the orchestrator uses a fine-tuned binary classifier (see the details in App. \ref{app:sec_binary_classifier}) to determine whether a query pertains to Islamic content. Queries predicted as Islamic are routed to \textit{our proposed multi-agent system}, while all other queries are handled by general-purpose assistants. To evaluate this classifier, we developed a dataset of 1{,}700 queries annotated by three independent annotators. Inter-annotator agreement is 0.753, measured using Cohen's $\kappa$. The classifier achieves a macro-F1 score of 93.40. Further details on the classifier and evaluation dataset are provided in App.~\ref{app:sec_binary_classifier}.

\noindent
\textbf{Real-world usage.} Through the chat interface and API, the system has been used $\approx$1.9M times, in less than a year, demonstrating its practical utility in real-world settings. 
In 6,441 queries user were provided rating in terms of like and dislike, in which 77.4\% cases users liked the responses.


\section{Conclusion}
In this paper, we present \sadiq{}, a tool-routed \textit{multi-agent architecture} for Islamic QA that supports heterogeneous user intents.
Unlike fixed \textit{retrieve-then-generate} pipelines, the system separates \textit{(i)} retrieval-grounded fiqh and general Islamic knowledge QA with traceable evidence, \textit{(ii)} canonical scripture handling where verbatim correctness is required, and \textit{(iii)} rule- and arithmetic-constrained obligations such as zakat and inheritance via deterministic computation and invariant checks. This design targets common failure modes in Islamic QA, including misquotation, weak attribution for jurisprudential claims, and numerically inconsistent calculations. Evaluations on public Islamic QA benchmarks show that combining intent routing, specialized tools, and post-generation verification can improve reliability in Islamic knowledge systems. Future work will expand jurisprudential coverage across schools of thought, improve routing robustness, 
and strengthen quotation validation for Hadith collections.



\section*{Limitations}
\label{sec:limitations}
The proposed system supports Islamic knowledge QA. However, it does not replace qualified scholarly authority or issue binding fatwas. Its responses depend on the coverage and quality of the retrieval corpora, curated sources, and routing decisions. Routing errors may send calculation-heavy queries to free-form fiqh QA or select a suboptimal module. Although we use citation tracking and verification, citations may remain incomplete, and retrieved evidence may reflect jurisprudential diversity that is difficult to summarize without oversimplification.
The deterministic calculators also have scope limits. Inheritance outcomes depend on correctly specified heirs and assumptions, and the implementation may cover only selected schools or disputed cases. Zakat, calendar, and prayer-time outputs depend on user-provided parameters, calculation methods, local conventions, and moon-sighting criteria. Finally, open-ended evaluation partly relies on automated or LLM-based judging, which may miss nuance, context, or legitimate scholarly disagreement.

\section*{Broader Impact}
\label{sec:broader_impact}
Our multi-agent Islamic QA system can broaden access to grounded information by helping users retrieve canonical references, navigate common questions, and perform rule-based computations such as zakat and inheritance with transparent outputs. This can support education, personal learning, and community use, especially in bilingual settings. However, the system also introduces risks. Users may over-trust outputs, overlook conditional rulings, or treat summarized answers as universally applicable despite legitimate differences across schools, locales, and circumstances. The system may also be misused for selective quotation, sectarian framing, or misleading claims. To mitigate these risks, our design emphasizes traceability through citations and audit traces, explicit handling of disagreement when relevant, scoped answers, uncertainty signaling, and recommendations to consult qualified scholars for high-stakes or personal matters. Deployment should also include privacy-preserving logging, data minimization, and continuous monitoring to reduce unintended harms.


\bibliography{bibliography/bibliography}

\appendix

\section*{Appendix}


\section{Islamic vs. Non-Islamic Classifier}
\label{app:sec_binary_classifier}
\noindent\textbf{Training data and model.}
We train a binary \emph{Islamic vs.\ non-Islamic} query classifier using a knowledge-distillation setup: a large teacher model produces offline labels indicating whether a query requires Islamic religious sources (e.g., Qur'an, Hadith, tafsir, fiqh), and a lightweight classifier is trained for low-latency inference. We curate $\sim$2.13M user queries annotated with binary labels (637,748 positive; 1,488,793 negative; ratio $\approx$1:2.3) spanning Arabic and English, with a median length of 52 characters. 

The model is implemented by adding a linear prediction head on top of a \texttt{bge-m3} encoder. We freeze the encoder and train only the head for 20 epochs with learning rate $3\times 10^{-4}$ and BF16 mixed precision, selecting the best checkpoint by macro-F1 on a stratified held-out validation split (10\%). At inference time, we binarise the continuous output using a threshold of 0.66. Negative coverage includes general reasoning and math-style queries sampled from publicly available sources such as LMSYS-Chat-1M~\cite{zheng2023lmsyschat1m}, WildChat~\cite{zhao2024wildchat}, and MetaMath~\cite{yu2024metamathbootstrapmathematicalquestions}.

\noindent\textbf{Evaluation dataset.} 
We developed a dataset of 1,716 queries. Each query was labeled independently by three annotators, who received instructions and training. Annotators were compensated at a standard hourly rate. Inter-annotator agreement, measured using Cohen's $\kappa$, was 0.753, with an overall label agreement of 88.2\%.

\noindent\textbf{Results.}
On this benchmark, the classifier achieves a precision of 0.922, recall of 0.924, F1 score of 0.923, and accuracy of 0.944 at a threshold of 0.66.

\section{Prompts}
\label{sec-app-prompts}

\subsection{Islamic Query Classifier}
\label{sec-app-prompts-question}

\begin{lstlisting}[style=pythonstyle,language=TeX,caption={Prompt for classifying Islamic questions into task categories.},label={lst:prompt_islamic_question_classification}]
You are an expert **Islamic question classifier**.

Analyze the user's question and classify it into **ONE** of these categories:

1. **fiqh_ruling**: Questions asking for Islamic legal rulings, permissibility, obligations, or jurisprudence
   Examples: "Is X halal?", "What's the ruling on Y?",
   (*@\<هل هذا حلال؟>@*), (*@\<ما حكم كذا؟>@*)

2. **quran_retrieval**: Questions asking for specific Quranic verses or ayahs
   Examples: "What does verse 2:255 say?", "Find ayah about patience",
   (*@\<ما هي الآية رقم 255 من سورة البقرة؟>@*), (*@\<اكتب الآية 275 من سورة البقرة>@*)

3. **general_islamic**: General questions about Islamic knowledge, history, concepts, or practices
   % Use this when the question does NOT request a ruling/calculation/timing/retrieval explicitly.
   Examples: "Who was Umar ibn al-Khattab?", "What is tawakkul?", (*@\<ما معنى الإحسان؟>@*)

4. **greeting**: Simple greetings, thanks, or pleasantries
   Examples: "Hi", "Thanks!", (*@\<السلام عليكم>@*), (*@\<جزاك الله خيرًا>@*)

5. **zakat_calculation**: Requests to compute Zakat owed based on assets, debts, or metal prices
   Examples: "How much zakat do I pay on $10,000?", (*@\<زكاة المال كم؟>@*)

6. **inheritance_calculation**: Requests to divide an estate among heirs (Mirath/Faraid)
   Examples: "Split inheritance among wife and children", (*@\<قسمة الميراث بين الورثة>@*)

7. **dua_lookup**: Requests for duas (supplications) or adhkar (remembrances), or what to say in specific situations
   Examples: "dua for entering bathroom", "morning adhkar", "what to say before sleeping",
   (*@\<دعاء دخول الحمام>@*)

8. **islamic_calendar**: Questions about Hijri/Islamic dates, date conversions, or Islamic events/holidays
   Examples: "What is today's Hijri date?", "When is Ramadan 2025?", "Convert March 1 to Hijri", "When is Eid?",
   (*@\<ما هو التاريخ الهجري اليوم؟>@*), (*@\<متى رمضان؟>@*)

9. **prayer_times**: Questions about prayer times, salah timing, or Qibla direction for a location
   Examples: "What time is Fajr in Dubai?", "Prayer times for London", "Which direction is Qibla from Tokyo?",
   (*@\<أوقات الصلاة في الرياض>@*), (*@\<اتجاه القبلة>@*)

Return ONLY valid JSON in this format (no markdown, no explanation):
{
  "question_type": "fiqh_ruling",
  "language": "en",
  "confidence": 0.95,
  "reasoning": "Brief explanation",
  "subquestions": ["question1"],
  "requires_retrieval": true
}

Classify the question below:

Question: {question}
\end{lstlisting}

\subsection{Quran Related Queries}
\label{sec-app-prompts-quran}

\begin{lstlisting}[style=pythonstyle,language=TeX,caption={Prompt for classifying Quran-related questions into sub-types.},label={lst:prompt_quran_subtype_classification}]
You are an expert at classifying **Quran-related questions**.

Classify the user's Quran question into **ONE** of these sub-types:

1. **specific_verse**: Asking for a specific verse by number or reference
   Examples:
   - "What does verse 2:255 say?"
   - "Show me ayah 7 of Al-Fatiha"
   - (*@\<اكتب الآية 275 من سورة البقرة>@*)
   - (*@\<ما هي آخر ثلاث آيات من سورة البقرة؟>@*)
   - "What are the last three verses of Surah Al-Baqarah?"

2. **full_surah**: Asking for an entire surah's text
   Examples:
   - "Write Surah Al-Fatiha"
   - (*@\<اكتب سورة الإخلاص>@*)
   - "Give me the entire Surah Nas"

3. **statistics**: Counting verses, surah metadata, or structural queries
   Examples:
   - "How many verses in Surah Al-Baqarah?"
   - (*@\<كم عدد الآيات في سورة الكهف؟>@*)
   - "Which surah has the most verses?"
   - "Is Al-Baqarah Makki or Madani?"
   - (*@\<كم عدد آيات سورة الفاتحة؟>@*)

4. **interpretation**: Asking for meaning, tafsir, or explanation
   Examples:
   - "What is the meaning of Ayat al-Kursi?"
   - (*@\<ما معنى آخر آيات سورة البقرة؟>@*)
   - "Explain the interpretation of Al-Kawthar"
   - "What does the Quran say about patience?"

Return ONLY the sub-type name (specific_verse, full_surah, statistics, or interpretation).

Question: {question}

Sub-type:
\end{lstlisting}

\subsection{Dua}
\label{sec-app-prompts-dua}

\begin{lstlisting}[style=pythonstyle,language=Python,caption={Bilingual (Arabic/English) prompt construction for matching du\textquotesingle a occasions.},label={lst:dua_prompt_construction}]
if lang == "ar":
    system_prompt = """(*@\<أنت مساعد متخصص في تحديد المناسبات
    المناسبة للأدعية الإسلامية.
مهمتك: حدد أرقام المناسبات التي 
تتوافق فعلاً مع سؤال المستخدم.
أجب فقط بالأرقام مفصولة بفواصل (مثال: 1,3)
إذا لم تجد أي مناسبة مطابقة، أجب بـ "none".>@*)"""
    user_prompt = f"""(*@\<سؤال المستخدم:>@*) {question}

(*@\<المناسبات المرشحة:>@*)
{occasions_list}

(*@\<ما هي أرقام المناسبات المطابقة لسؤال المستخدم؟>@*)"""
else:
    system_prompt = """You are a specialist in matching Islamic dua occasions to user queries.
Your task: Identify which occasion numbers actually match the user's question.
Respond ONLY with comma-separated numbers (e.g., 1,3)
If no occasions match, respond with "none"."""
    user_prompt = f"""User's question: {question}

Candidate occasions:
{occasions_list}

Which occasion numbers match the user's question?"""
\end{lstlisting}

\subsection{LLM-as-a-Judge}
\label{app:grader_template}

\begin{lstlisting}
Your job is to look at a question, a gold target, and a predicted answer, and then assign a grade of either ["CORRECT", "INCORRECT", "NOT_ATTEMPTED"].
First, I will give examples of each grade, and then you will grade a new example.


The following are examples of CORRECT predicted answers.


Question: What are the names of Barack Obama's children?
Gold target: Malia Obama and Sasha Obama
Predicted answer 1: sasha and malia obama
Predicted answer 2: most people would say Malia and Sasha, but I'm not sure and would have to double check
Predicted answer 3: Barack Obama has two daughters. Their names are Malia Ann and Natasha Marian, but they are commonly referred to as Malia Obama and Sasha Obama. Malia was born on July 4, 1998, and Sasha was born on June 10, 2001.


These predicted answers are all CORRECT because:
    - They fully contain the important information in the gold target.
    - They do not contain any information that contradicts the gold target.
    - Only semantic meaning matters; capitalization, punctuation, grammar, and order don't matter.
    - Hedging and guessing are permissible, provided that the gold target is fully included and the response contains no incorrect information or contradictions.


The following are examples of INCORRECT predicted answers.


Question: What are the names of Barack Obama's children?
Gold target: Malia and Sasha
Predicted answer 1: Malia.
Predicted answer 2: Malia, Sasha, and Susan.
Predicted answer 3: Barack Obama does not have any children.
Predicted answer 4: I think it's either Malia and Sasha. Or it could be Malia and Jackie. Or it could be Joey and Malia.
Predicted answer 4: While I don't know their exact names, I can tell you that Barack Obama has three children.
Predicted answer 5: It's possible you may mean Betsy and Olivia. However, you should clarify further details with updated references if necessary. Is that the correct answer?
Predicted answer 6: It may be the case that Obama's child is named James. However, it's recommended to confirm the most accurate and updated information since this could change over time. This model may not always reflect the most current information.


These predicted answers are all INCORRECT because:
    - A factual statement in the answer contradicts the gold target. Incorrect statements that have some hedging (e.g., "it is possible that", "although i'm not sure, i think") are also considered incorrect.


The following are examples of NOT_ATTEMPTED predicted answers.


Question: What are the names of Barack Obama's children?
Gold target: Malia and Sasha
Predicted answer 1: I don't know.
Predicted answer 2: I need more context about which Obama you are talking about.
Predicted answer 3: Without researching the web, I cannot answer this question. However, I can tell you that Barack Obama has two children.
Predicted answer 4: Barack Obama has two children. I know that one of them is Malia, but I'm not sure about the other one.


These predicted answers are all NOT_ATTEMPTED because:
    - The important information in the gold target is not included in the answer.
    - No statements in the answer contradict the gold target.


Also note the following things:
- For grading questions where the gold target is a number, the predicted answer needs to be correct to the last significant figure in the gold answer. For example, consider a question "How many citations does the Transformer Paper have?" with gold target "120k". 
    - Predicted answers "120k", "124k", and 115k" are all CORRECT. 
    - Predicted answers "100k" and "113k" are INCORRECT. 
    - Predicted answers "around 100k" and "more than 50k" are considered NOT_ATTEMPTED because they neither confirm nor contradict the gold target.
- The gold target may contain more information than the question. In such cases, the predicted answer only needs to contain the information that is in the question.
    - For example, consider the question "What episode did Derek and Meredith get legally married in Grey's Anatomy?" with gold target "Season 7, Episode 20: White Wedding". Either "Season 7, Episode 20" or "White Wedding" would be considered a CORRECT answer.
- Do not punish predicted answers if they omit information that would be clearly inferred from the question.
    - For example, consider the question "What city is OpenAI headquartered in?" and the gold target "San Francisco, California". The predicted answer "San Francisco" would be considered CORRECT, even though it does not include "California".
    - Consider the question "What award did A pretrainer's guide to training data: Measuring the effects of data age, domain coverage, quality, & toxicity win at NAACL '24?", the gold target is "Outstanding Paper Award". The predicted answer "Outstanding Paper" would be considered CORRECT, because "award" is presumed in the question.
    - For the question "What is the height of Jason Wei in meters?", the gold target is "1.73 m". The predicted answer "1.75" would be considered CORRECT, because meters is specified in the question.
    - For the question "What is the name of Barack Obama's wife?", the gold target is "Michelle Obama". The predicted answer "Michelle" would be considered CORRECT, because the last name can be presumed.
- Do not punish for typos in people's name if it's clearly the same name. 
    - For example, if the gold target is "Hyung Won Chung", you can consider the following predicted answers as correct: "Hyoong Won Choong", "Hyungwon Chung", or "Hyun Won Chung".


Here is a new example. Simply reply with either CORRECT, INCORRECT, NOT ATTEMPTED. Don't apologize or correct yourself if there was a mistake; we are just trying to grade the answer.

Question: question
Gold target: target
Predicted answer: predicted_answer

Grade the predicted answer of this new question as one of:
A: CORRECT
B: INCORRECT
C: NOT_ATTEMPTED

Just return the letters "A", "B", or "C", with no text around it.
\end{lstlisting}

\section{System Implementation Details}
\label{app:tool_details}


This section provides implementation details for all specialized tools and components in our multi-agent architecture described in Section~\ref{sec:system_design}. We present the design decisions, algorithms, and configuration strategies that enable robust Islamic QA across heterogeneous query types.

\subsection{Hybrid Query Classifier}
\label{app:classifier}

The hybrid query classifier serves as the system's entry point, performing deterministic routing based on predicted intent labels. Our classifier employs a two-tier approach combining LLM-based classification with a prototype-based fallback mechanism to ensure robust routing even when the primary classifier fails.

\subsubsection{LLM-Based Primary Classification}

The primary classifier prompts an LLM with structured instructions. The model outputs a JSON object with six fields. These include \textit{(i)} an intent label from nine predefined categories, \textit{(ii)} the detected language, \textit{(iii)} a confidence score between 0 and 1, \textit{(iv)} a brief rationale, \textit{(v)} optional decomposition into subquestions, and \textit{(vi)} a boolean flag indicating whether document retrieval is required. 

The intent categories are \textit{fiqh\_ruling}, \textit{quran\_retrieval}, \textit{general\_islamic}, \textit{greeting}, \textit{zakat\_calculation}, \textit{inheritance\_calculation}, \textit{dua\_lookup}, \textit{islamic\_calendar}, and \textit{prayer\_times}.

The classifier prompt (Listing~\ref{lst:prompt_islamic_question_classification}) provides explicit examples for each category in both Arabic and English to ensure consistent classification across languages. This bilingual exemplar approach is crucial for handling code-switching and dialectal variation common in user queries.

The LLM operates at zero temperature to maximize determinism and outputs strictly formatted JSON. We strip common model artifacts including end-of-turn tokens and extract JSON from markdown code blocks when models wrap their output. The classification temperature is configurable through a three-tier settings hierarchy. However, it defaults to 0.0, with a maximum output length of 300 tokens to encourage concise rationales.

\subsubsection{Embedding-Based Fallback Mechanism}

When LLM classification fails due to low confidence (below 0.5), malformed JSON output, or exception during invocation, the system seamlessly falls back to an embedding-based classifier. This fallback mechanism computes cosine similarity between the query embedding and pre-computed prototype embeddings for each intent-language pair. The confidence score is derived from the margin between the top two similarity scores using the formula $\text{confidence} = \frac{\text{sim}_1 - \text{sim}_2}{2} + 0.5$, which maps the separation between candidates to a 0-1 range. Class-specific rules then determine the retrieval flag based on the predicted intent.

In Table~\ref{tab:prototype_examples}, we present examples for each language-intent pair. We pre-compute their embeddings offline using Qwen3-Embedding-4B and cache them in memory for efficient lookup. This dual-tier design is important for multi-tool Islamic QA because routing errors can be costly. For example, sending an inheritance query to a generative fiqh agent may produce outputs that deviate from arithmetic or legal constraints.


\subsubsection{Evaluation} 
\label{ssec_app_query_classifier}
To evaluate the \textit{hybrid query classifier}, we developed an intent-labeled dataset of 705 real user queries sampled from the system's chat interface. We anonymized the queries and removed personally identifying information before annotation. Six annotators labeled each query into one of nine intent categories, with three independent labels per query. We assigned the final label by majority vote and discarded instances without majority agreement. The final distribution is \texttt{fiqh\_ruling} (31.4\%), \texttt{general\_islamic} (29.1\%), \texttt{inheritance\_calculation} (17.4\%), \texttt{zakat\_calculation} (5.3\%), \texttt{quran\_retrieval} (4.7\%), \texttt{dua\_lookup} (3.9\%), \texttt{islamic\_calendar} (3.6\%), \texttt{prayer\_times} (2.4\%), and \texttt{greeting} (2.1\%).
We measure inter-annotator agreement with Fleiss' $\kappa$, obtaining $\kappa = 0.76$ across the three annotations, indicating substantial agreement.

We use this dataset to benchmark routing performance and compare against strong LLM-only baselines. As presented in Table \ref{tab:classifier_eval}, our hybrid classifier achieves 90.1\% accuracy, while zero-shot GPT-5 and Gemini achieve 89.3\% and 89.7\% accuracy, respectively.



\begin{table}[t]
\centering
\setlength{\tabcolsep}{3pt}
\scalebox{0.85}{
\begin{tabular}{lc}
\toprule
\textbf{Classifier} & \textbf{Accuracy (\%)} \\
\midrule
Ours (Hybrid) & 90.1 \\
GPT-5 (zero-shot) & 89.3 \\
Gemini (zero-shot) & 89.7 \\
\bottomrule
\end{tabular}
}
\vspace{-0.2cm}
\caption{Hybrid query classifier classification accuracy.
}
\label{tab:classifier_eval}
\vspace{-0.3cm}
\end{table}



\subsubsection{Examples: Query Types}
In Table~\ref{tab:prototype_examples}, we present the query types along with bilingual examples.

\begin{table*}[t]
\centering
\setlength{\tabcolsep}{2pt}
\scalebox{0.8}{%
\begin{tabular}{p{0.25\linewidth} p{0.43\linewidth} p{0.30\linewidth}}
\toprule
\textbf{Type} & \textbf{English} & \textbf{Arabic} \\
\midrule
Fiqh ruling &
What is the ruling on music in Islam? &
\AR{\smallما حكم الموسيقى في الإسلام؟} \\
Quran retrieval &
Quote Surah Al-Baqarah verse 275. &
\AR{\smallاكتب الآية 275 من سورة البقرة} \\
General islamic &
What are the five pillars of Islam? &
\AR{\smallما هي أركان الإسلام الخمسة؟} \\
Greeting &
Assalamu alaikum. &
\AR{\smallالسلام عليكم} \\
Zakat calculation &
I have 100 grams of gold, how much zakat? &
\AR{\smallاحسب زكاتي على الذهب} \\
Inheritance calculation &
What is the share of wife in inheritance? &
\AR{\smallما نصيب الزوجة من الميراث؟} \\
Dua lookup &
What is the dua for entering the toilet? &
\AR{\smallما هو دعاء دخول الحمام؟} \\
Islamic calendar &
What is today's Hijri date? &
\AR{\smallما هو التاريخ الهجري اليوم؟} \\
Prayer times &
What time is Fajr in Dubai? &
\AR{\smallمتى صلاة الفجر في دبي؟} \\ 
Quran statistics &
How many verses in Surah Al-Baqarah? &
\AR{\smallكم عدد آيات سورة البقرة؟} \\ 
Quran interpretation &
What is the meaning of Ayat al-Kursi? &
\AR{\smallما معنى آية الكرسي؟} \\
\bottomrule
\end{tabular}
}
\vspace{-0.2cm}
\caption{Representative English--Arabic query-type examples.}
\label{tab:prototype_examples}
\vspace{-0.3cm}
\end{table*}
\subsection{Greeting Tool}
\label{app:greeting_tool}

The greeting tool handles simple greetings and pleasantries with culturally appropriate Islamic responses. Language detection operates on Arabic character ratio, classifying text as Arabic when more than 30\% of characters fall in the Unicode Arabic blocks (U+0600–U+06FF). 

For Arabic queries, the system responds in MSA with traditional Islamic greetings and maintains formal register. For English queries, responses include transliterated Arabic phrases such as ``Wa alaykum assalam wa rahmatullahi wa barakatuh'' followed by an offer to assist with Islamic knowledge. All responses are constrained to one or two sentences to maintain brevity while conveying warmth.

The tool operates with configurable temperature (default 0.2) and maximum token length (default 256). If LLM invocation fails, the system returns language-appropriate fallback greetings:

\AR{\small وعليكم السلام ورحمة الله وبركاته. كيف يمكنني مساعدتك في أمور الإسلام؟} 

\noindent for Arabic, and \textit{``Wa alaykum assalam wa rahmatullahi wa barakatuh. How may I assist you with Islamic knowledge today?''} for English.

\subsection{Islamic Calendar Tool}
\label{app:islamic_calendar}

The Islamic calendar tool handles Hijri date queries, conversions, and Islamic event lookups through deterministic rule-based processing. Query type detection operates via multilingual keyword matching to classify inputs into five subtypes such as \textit{(i)} current Hijri date, \textit{(ii)} Gregorian-to-Hijri conversion, \textit{(iii)} Hijri-to-Gregorian conversion, \textit{(iv)} specific Islamic event dates, and \textit{(v)} upcoming events listing.

Date conversions rely on the \texttt{hijri-converter} library,\footnote{\url{https://pypi.org/project/hijridate/}}, which implements the Umm al-Qura calendar system. All conversions account for three critical factors, including \textit{(i)} lunar month visibility rules based on astronomical calculations, \textit{(ii)} regional variation in moon sighting practices (observational versus calculated calendars), and \textit{(iiii)} the distinction between arithmetic approximation and actual visibility. The system includes explicit disclaimers regarding local moon-sighting variations, acknowledging that Islamic calendar dates may differ by one day based on regional authorities.

Event resolution operates over a curated bilingual ontology containing 20+ major Islamic events with English and Arabic names, precise Hijri month and day-of-month specifications, and event type classifications distinguishing religious obligations from recommended practices and commemorative dates. The system implements year-rollover logic. When an event has already occurred in the current Islamic year, it returns the next occurrence in the following year. Major events include Ramadan beginning (Ramadan 1), Eid al-Fitr (Shawwal 1), Day of Arafah (Dhul-Hijjah 9), Eid al-Adha (Dhul-Hijjah 10), and Ashura (Muharram 10).
Output formatting adapts to language, using Arabic-Indic numerals 
%
%
\noindent for Arabic responses and Western numerals for English. Each response includes the Hijri date with full month name, Gregorian equivalent, localized day of week, and a disclaimer noting that actual dates depend on local moon sighting and may vary by region.

\subsection{Prayer Times and Qibla Tool}
\label{app:prayer_times}

This tool computes Islamic prayer times and Qibla direction using astronomical calculations based on geographic coordinates. Location resolution employs a \textit{four-stage pipeline} designed for accuracy and robustness. 
\textit{First}, the system attempts exact or fuzzy matching against a curated database of over 8,000 cities with pre-computed coordinates, time zones, and preferred calculation methods. 
\textit{Second}, if database lookup fails, an LLM extraction step parses city names from natural language at zero temperature, including transliteration from Arabic or other languages to English. 
\textit{Third}, when LLM extraction yields no result, the system falls back to a rate-limited external geocoding API. 
\textit{Finally}, if all resolution methods fail, the system defaults to Doha, Qatar (25.2854°N, 51.5310°E) with an explicit disclaimer.

Prayer time calculation employs the \texttt{pyIslam} library\footnote{\url{https://pypi.org/project/islam/}} with support for four internationally recognized calculation methods, each defined by specific angular parameters for Fajr (pre-dawn) and Isha (night) prayers. Table~\ref{tab:prayer_methods} presents these methods with their respective angles.

\begin{table}[h]
\centering
\setlength{\tabcolsep}{2pt}
\scalebox{0.7}{
\begin{tabular}{lrr}
\toprule
\textbf{Method} & \textbf{Fajr Angle} & \textbf{Isha Angle} \\
\midrule
Muslim World League & 18° & 17° \\
Egyptian Authority & 19.5° & 17.5° \\
Umm al-Qura (Makkah) & 18.5° & 90 min after Maghrib \\
Islamic Society of North America & 15° & 15° \\
\bottomrule
\end{tabular}}
\vspace{-0.2cm}
\caption{Prayer time calculation methods and their angular parameters. The Fajr angle determines when morning twilight begins, while the Isha angle marks when evening twilight ends.}
\label{tab:prayer_methods}
\vspace{-0.3cm}
\end{table}

The calculation requires four inputs. \textit{(i)} latitude and longitude in decimal degrees, \textit{(ii)} UTC offset for the location's time zone, \textit{(iii)} the selected calculation method (defaulting to Muslim World League), \textit{(iv)} and the target date (defaulting to the current day). Output includes precise times for all five daily prayers, \textit{Fajr}, \textit{Dhuhr}, \textit{Asr}, \textit{Maghrib}, and \textit{Isha}, with sunrise and astronomical midnight.

Qibla direction calculation determines the great-circle bearing to the Kaaba in Makkah (21.4225°N, 39.8262°E) using the spherical geometry formula:
$$\theta = \arctan\left(\frac{\sin(\Delta\lambda)}{\cos(\phi_1)\tan(\phi_2) - \sin(\phi_1)\cos(\Delta\lambda)}\right)$$
where $\phi_1, \lambda_1$ represent the current location's coordinates, $\phi_2, \lambda_2$ are Makkah's coordinates, and $\Delta\lambda = \lambda_2 - \lambda_1$ is the longitude difference. The bearing $\theta$ is then converted to a compass heading (0-360°) and mapped to cardinal directions. The system also computes and reports the great-circle distance to Makkah in kilometers.

\subsection{Dua Lookup Tool}
\label{app:dua_lookup}

The Dua lookup tool provides verbatim retrieval of authenticated Islamic supplications from curated sources, designed explicitly to prevent generative hallucination. The tool employs a \textit{two-stage retrieval-and-selection architecture} that separates semantic matching from relevance filtering.

In the \textit{first stage}, the system computes a query embedding using Qwen3-Embedding-4B and ranks pre-computed occasion embeddings by cosine similarity. We retain the top-$k$ candidates (default $k=5$) with minimum similarity threshold 0.2, producing structured candidates containing the internal page title identifier, English occasion description, and similarity score.

The \textit{second stage} employs a lightweight LLM as a precision filter. The prompt (Listing~\ref{lst:dua_prompt_construction}) presents numbered occasion candidates and instructs the LLM to output comma-separated indices of relevant occasions. Selected indices are deterministically mapped to page title keys for exact retrieval.

Each Dua record contains eight fields, \textit{(i)} optional title, \textit{(ii)} diacritized Arabic text, \textit{(iii)} English translation, \textit{(iv)} primary source reference, e.g., ``Sahih Bukhari 6306'', \textit{(v)} canonical reference URL, \textit{(vi)} page title identifier, \textit{(vii)} English occasion description, and \textit{(viii)} Arabic occasion description. The tool returns these records verbatim without generative rewriting. This preserves the authenticity of Arabic text, source attribution, diacritics, and consistency across queries.


\subsection{Zakat Calculator}
\label{app:zakat_calculator}

The Zakat calculator performs deterministic Shariah-compliant computation for the obligatory 2.5\% annual levy on eligible wealth. It supports five asset groups:
\begin{enumerate}[noitemsep,topsep=0em,leftmargin=1.5em,labelsep=.5em]
    \item \textbf{Monetary assets}, including cash, gold, silver, business inventory, stocks, and receivable debts, using a 2.5\% rate.
    \item \textbf{Agricultural produce}, using irrigation-dependent rates of 10\% for rain-fed crops, 5\% for irrigated crops, and 7.5\% for mixed irrigation.
    \item \textbf{Livestock}, following Hadith-based schedules for camels, cattle, and sheep.
    \item \textbf{Foreign currency}, supporting conversion across 15+ denominations.
    \item \textbf{Major cryptocurrencies}, including Bitcoin and Ethereum.
\end{enumerate}


%

\noindent\textbf{Nisab computation} determines the minimum wealth threshold for Zakat obligation:
\[
\text{Nisab} = \min(85\text{ g} \times P_{\text{gold}}, 595\text{ g} \times P_{\text{silver}})
\]
We use the lower threshold, following the scholarly view that this is more beneficial to the poor by broadening eligibility.

Algorithm~\ref{alg:zakat_calc} summarizes the calculation procedure. It aggregates monetary assets, computes net wealth after deductions, applies the 2.5\% rate when assets exceed Nisab, checks agricultural produce against the 653 kg threshold, applies livestock lookup tables, and sums all categories. 
Validation enforces strict checks. Inputs must be non-negative and finite, gold prices must exceed silver prices, and produce weights must meet required thresholds. Outputs include total Zakat due, category-level breakdowns, Nisab information, and warnings about holding periods and price verification.

\begin{algorithm}[!tbh]
\caption{Zakat Calculation}
\label{alg:zakat_calc}
\begin{algorithmic}[1]
\REQUIRE Assets $A$, Liabilities $L$, Prices $P$

\STATE $N_{\text{gold}} \leftarrow 85 \times P_{\text{gold/gram}}$
\STATE $N_{\text{silver}} \leftarrow 595 \times P_{\text{silver/gram}}$
\STATE $N \leftarrow \min(N_{\text{gold}}, N_{\text{silver}})$

\STATE $A_{\text{monetary}} \leftarrow A_{\text{cash}}
    + (A_{\text{gold}} \times P_{\text{gold}})
    + (A_{\text{silver}} \times P_{\text{silver}})
    + A_{\text{business}} + A_{\text{stocks}}$

\STATE $A_{\text{net}} \leftarrow A_{\text{monetary}} - L_{\text{debts}}$

\IF{$A_{\text{net}} \ge N$}
    \STATE $Z_{\text{monetary}} \leftarrow 0.025 \times A_{\text{net}}$
\ELSE
    \STATE $Z_{\text{monetary}} \leftarrow 0$
\ENDIF

\STATE $Z_{\text{agriculture}} \leftarrow \text{AgricultureZakat}(A_{\text{produce}})$
\STATE $Z_{\text{livestock}} \leftarrow \text{LivestockZakat}(A_{\text{livestock}})$

\STATE $Z \leftarrow Z_{\text{monetary}} + Z_{\text{agriculture}} + Z_{\text{livestock}}$
\ENSURE $Z$ with category breakdown. 
\end{algorithmic}
\vspace{-0.1cm}
\end{algorithm}


\subsection{Inheritance Calculator}
\label{app:inheritance_calculator}

\begin{figure}[!tbh]
    \centering
    \includegraphics[width=0.9\linewidth]{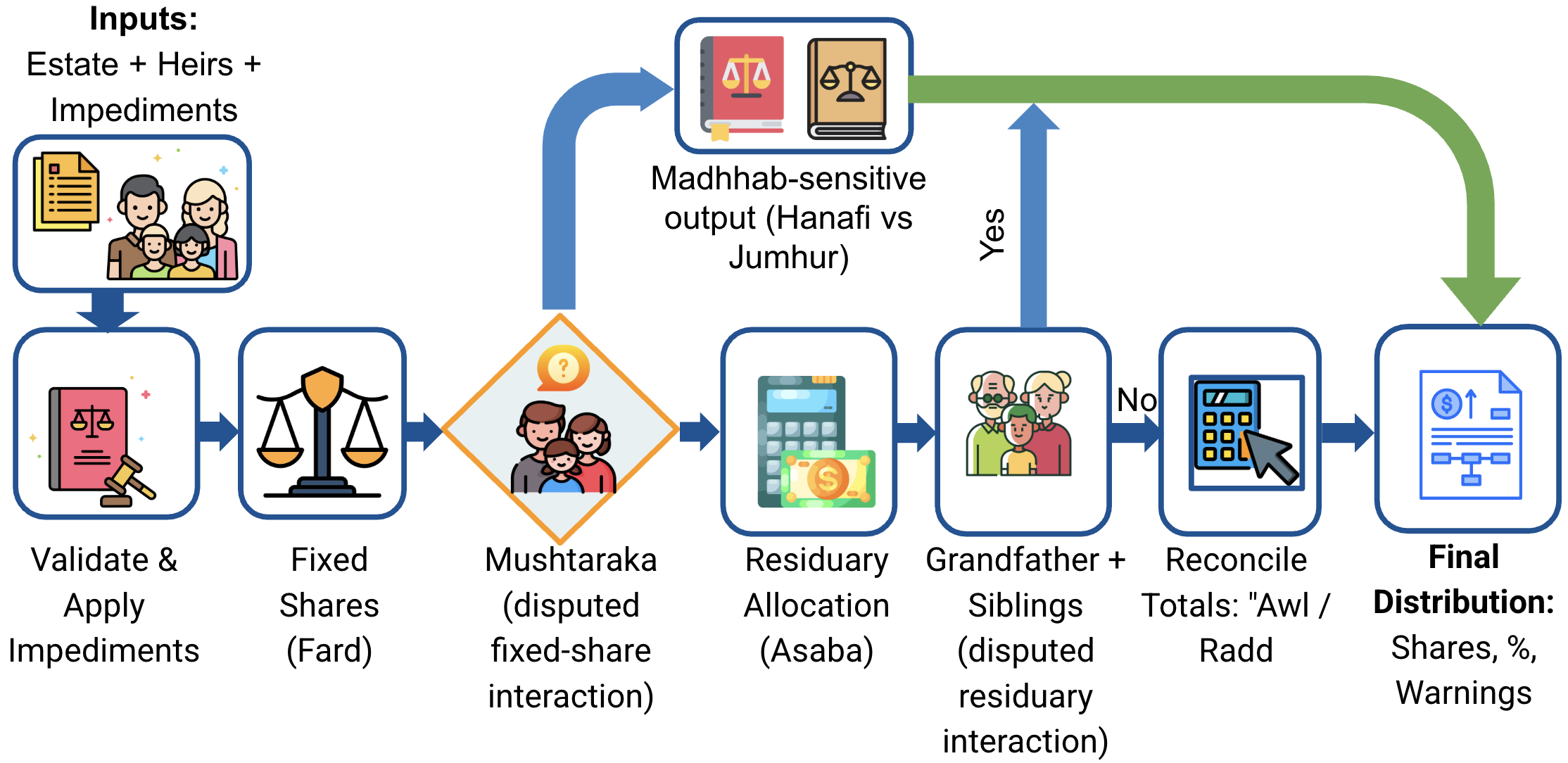}
    \vspace{-0.3cm}
    \caption{Inheritance calculation workflow. Disputed cases return parallel outcomes instead of collapsing to a single ruling.}
    \vspace{-0.3cm}
    \label{fig:fanar_sadiq_inheritance_system}
\end{figure}

The inheritance calculator implements Islamic Faraid using Quranic rules with explicit madhhab handling. In Figure \ref{fig:fanar_sadiq_inheritance_system}, we present the inheritance calculation workflow. The calculator proceeds through \textit{three phases} mirroring classical methodology.

\textbf{\textit{Phase one}} assigns fixed shares (Fard) according to Quranic specifications. Table~\ref{tab:inheritance_fard} presents representative allocations covering husbands, wives, parents, and children under various conditions.

\begin{table}[h]
\centering
\scalebox{0.75}{
\begin{tabular}{lll}
\toprule
\textbf{Heir} & \textbf{Conditions} & \textbf{Share} \\
\midrule
Husband & No children & 1/2 \\
Husband & Has children & 1/4 \\
Wife & No children & 1/4 \\
Wife & Has children & 1/8 \\
Father & Has children/grandchildren & 1/6 \\
Mother & Has children/grandchildren & 1/6 \\
Daughter (sole) & No sons & 1/2 \\
Daughters ($\geq$2) & No sons & 2/3 \\
\bottomrule
\end{tabular}
}
\vspace{-0.2cm}
\caption{Representative fixed share (Fard) allocations from Quranic specifications.}
\label{tab:inheritance_fard}
\vspace{-0.3cm}
\end{table}

\textbf{\textit{Phase two}} distributes remaining estate through the Asaba system following a strict priority chain: sons and daughters (2:1 ratio), grandsons through sons, father, grandfather, brothers (full then paternal), nephews, uncles, and cousins.

\textbf{\textit{Phase three}} enforces arithmetic consistency through \textit{Awl}, which proportionally reduces shares when they exceed 100\%, and \textit{Radd}, which redistributes surplus shares. When madhhab differences apply, such as Radd spouse eligibility, grandfather-sibling competition, or cases like Mushtaraka, the calculator returns parallel school-specific distributions rather than collapsing them into a single ruling.
Input schema accommodates 27 parameters covering all heir relationships. Output provides individual allocations, school difference flags, parallel distributions when schools diverge, Awl/Radd metadata, and warnings about special cases.

\subsubsection{Benchmark for Inheritance Cases}
To evaluate the inheritance calculator against real jurisprudential distributions, we construct a focused benchmark by automatically crawling the Arabic inheritance-fatwa section of IslamWeb and traversing its paginated listing pages. Each candidate fatwa link is canonicalized and deduplicated, then the full page is fetched and cleaned to remove navigation/template text. Content is segmented into title, question, and answer using Arabic section markers, with explicit stop markers to avoid including unrelated footer material. We normalize Arabic orthographic variants, diacritics, and Arabic-Indic digits, and filter samples to ensure a genuine inheritance (\textit{mirath}) context.

Each fatwa is transformed into a structured benchmark record with two components: \textit{(i)} calculator inputs extracted from the question (e.g., deceased gender, estate value, and heir counts across supported heir classes), and \textit{(ii)} expected inheritance outcomes extracted from the answer. The answer parser prioritizes inheritance tables when present (deriving denominators and per-heir shares), and falls back to textual extraction of fractions and residuary phrasing when tables are absent. The pipeline records detailed quality flags (e.g., missing gender, unsupported heirs, ambiguous grandmother type, unexpected fractions, \textit{`awl}/scaling indicators), supports optional LLM-assisted extraction as a fallback enhancer, and writes successful and failed/partial extractions to separate outputs for auditability. In the current setup, this process yielded 887 successful benchmark records and 12 failed/partial records.

\subsubsection{Evaluation}
We compare calculator outputs against fatwa-stated distributions using semantic validation with an LLM-as-judge (GPT-5). In the \textit{primary run}, 883 cases were judged successfully: 802 were labeled \textit{match} (90.83\%), 35 \textit{mismatch} (3.96\%), 20 \textit{partial} (2.27\%), and 26 \textit{unclear} (2.94\%).
A \textit{second run} with an alternative GPT-5-based extraction/judging setup yielded 747 \textit{match} out of 884 judged cases (84.50\%), with mean confidence 0.9539. Overall, these results indicate high practical validity of the inheritance calculator while producing a focused set of disagreement cases for targeted error analysis and iterative improvement.

\subsection{Fiqh Reasoning Tool}
\label{app:fiqh_tool}


The Fiqh reasoning tool implements Usul al-Fiqh methodology through a \textit{three-stage pipeline}, \textit{(i)} document retrieval, \textit{(ii)} structured reasoning, and \textit{(iii)} citation extraction.

\textbf{\textit{Document retrieval}} performs semantic search over 50,000+ documents from Quran, authenticated Hadith, classical jurisprudential texts, contemporary fatwas, and scholarly articles. The retriever uses Qwen3-Embedding-4B embeddings with cosine similarity, optional cross-encoder reranking, source diversity enforcement, and metadata filtering. Default retrieval limit is 12 documents.

The \textbf{\textit{reasoning stage}} employs an LLM with a specialized Usul al-Fiqh system prompt that instructs the model to: state ruling scope explicitly, separate rulings from evidence, apply source hierarchy (Quran $>$ Sahih Hadith $>$ Consensus $>$ Analogy), assign citation tags using \texttt{[CITE:N]} format, invoke \texttt{get\_quran\_ayah} for exact verses, and acknowledge uncertainty when sources conflict.
Tool integration allows invoking \texttt{get\_quran\_ayah} during generation, returning verbatim verse text to prevent paraphrasing. 

\textbf{\textit{Citation extraction}} operates post-generation by parsing all \texttt{[CITE:N]} tags, mapping tags to source documents, building structured citation lists, and enriching sources with \texttt{was\_cited} flags.

The tool supports token-by-token streaming via Server-Sent Events with four event types: \texttt{delta} (tokens), \texttt{citations} (sources), \texttt{metadata} (analysis), and \texttt{done} (completion). Bilingual formatting ensures Arabic responses use right-to-left flow with Arabic-Indic numerals and formal register, while English responses use accessible terminology. Configuration includes temperature (default 0.1), maximum tokens (default 4500), and retrieval limit (default 12).

\subsection{Quran Retrieval Tool}
\label{app:quran_tool}


The Quran retrieval tool provides verbatim verse lookup with support for diverse reference formats, including numeric references such as \texttt{2:275}, named references such as \texttt{Al-Baqarah 275}, verbose references such as \texttt{Surah 2 Verse 275}, and fuzzy references such as \texttt{last 3 verses of Al-Baqarah}.
%

Surah name resolution first applies exact lookup over 114 standard names, with case-insensitive English matching and prefix stripping. If exact matching fails, it uses fuzzy matching based on Levenshtein distance and embedding similarity, and returns matches above a 0.6 confidence threshold.


The SQLite database contains unique verse identifiers, surah and ayah numbers, Uthmanic \texttt{AyahText} with diacritics, simplified Arabic for search, English translation, and structural metadata such as Juz, Hizb, Manzil, and Ruku.

Algorithm~\ref{alg:quran_retrieval} presents the procedure. It parses the reference, resolves the surah, validates the ayah range, fetches the verses, builds citation URLs using the pattern \texttt{https://quran.com/<surah>/<ayah>}, and formats the response.

Error handling provides specific messages for invalid surah identifiers, ayah numbers out of range, database failures, and malformed formats.

\begin{algorithm}[t]
\caption{Quran Verse Retrieval}
\label{alg:quran_retrieval}
\begin{algorithmic}[1]
\REQUIRE Query $q$ (reference string)
\STATE $(s, a_{\text{start}}, a_{\text{end}}) \leftarrow \text{ParseReference}(q)$
\STATE $s_{\text{num}} \leftarrow \text{ResolveSurah}(s)$
\IF{$s_{\text{num}} = \text{null}$ OR $a_{\text{start}}, a_{\text{end}}$ invalid}
    \ENSURE \texttt{Error with guidance}
\ENDIF
\STATE $V \leftarrow \text{QueryDatabase}(s_{\text{num}}, a_{\text{start}}, a_{\text{end}})$
\STATE $\text{url} \leftarrow \text{BuildCitationURL}(s_{\text{num}}, a_{\text{start}}, a_{\text{end}})$
\ENSURE $\text{FormatResponse}(V, \text{url})$
\end{algorithmic}
\vspace{-0.1cm}
\end{algorithm}

\subsection{NL2SQL Tool for Quran Queries}
\label{app:nl2sql_tool}

The NL2SQL tool translates natural language questions about Quran structure into SQL queries. It uses a three-tier fallback strategy, \textit{(i)} a specialized NL2SQL model, \textit{(ii)} a general-purpose LLM, and \textit{(iii)} traditional retrieval when SQL generation fails.

The primary NL2SQL model is guided by few-shot retrieval. The system embeds the input query, retrieves the top 5 similar examples from more than 200 curated pairs, and injects both the examples and database schema into the prompt. The example bank covers aggregation, filtering, text search, and range queries. Schema injection, shown in Listing~\ref{lst:quran_schema}, provides structural information with inline semantic comments.

Before execution, the generated SQL passes through normalization and validation. Arabic text correction applies NFD-to-NFC normalization, mojibake detection, and diacritic restoration. SQL validation checks syntax, injection patterns, column names, and query complexity. Valid queries are executed with read-only access, a 5-second timeout, and a 1000-row limit.

After execution, the tool enriches and formats the results. Count-only outputs are augmented with representative examples when useful. An LLM formatter converts raw results into natural language, while skipping outputs longer than 8000 characters. The tool uses conservative defaults, including temperature 0.1, three retry attempts, and a 30-second timeout.

\begin{lstlisting}[language=SQL,caption={Quran database schema},label={lst:quran_schema}]
CREATE TABLE Quran (
    ID INTEGER PRIMARY KEY,
    Surah INTEGER,           -- Surah number (1-114)
    Ayah INTEGER,            -- Ayah within surah
    AyahText TEXT,           -- Arabic (Uthmanic)
    SimpleText TEXT,         -- No diacritics
    Translation TEXT,        -- English
    Juz INTEGER,             -- Division (1-30)
    Revelation TEXT          -- 'Meccan'/'Medinan'
);
\end{lstlisting}

\subsubsection{Model Training}
\label{app:nl2sql_training}

\paragraph{Training data (48k NL2SQL pairs).}
The specialized NL2SQL model was trained on a 48k-example dataset constructed via a template-driven procedure. We authored a library of natural-language query templates spanning Quranic retrieval and analytics (e.g., ``Give me ayah [x] for surah [y]'', verse-range retrieval, juz-based filtering, revelation-class filtering, and counting/aggregation). Each natural-language template is paired with a corresponding parameterized SQL template over the schema in Listing~\ref{lst:quran_schema}. We then instantiated template variables with valid Quran values (surah numbers, ayah indices, juz IDs, and revelation labels) to produce aligned NL/SQL pairs. To increase robustness to real user input, we augmented $\approx$10\% of the natural-language queries with typos and spelling variants (generated automatically), while keeping the SQL target unchanged.

\paragraph{Fine-tuning setup (LoRA SFT).}
We fine-tuned \texttt{Qwen/Qwen3-4B-Instruct-2507} using supervised fine-tuning (SFT) with LoRA adapters.
Training used a maximum sequence length of 4096 tokens, 3 epochs, learning rate $1\times10^{-5}$ with a cosine scheduler and 0.1 warmup ratio, bf16 precision, and DeepSpeed ZeRO-3. LoRA was applied to all target modules with rank 8, alpha 16, and dropout 0.05. We used per-device batch size 1 with gradient accumulation of 16 steps, and saved checkpoints at each epoch (keeping the latest).

\subsubsection{Evaluation Dataset and Metric}
\label{app:nl2sql_eval_dataset}

To evaluate the system under realistic user distributions, we sampled natural-language queries from \emph{Fanar} usage logs and manually curated them into two benchmark subsets:
\begin{itemize}[noitemsep,topsep=0em,leftmargin=1.5em,labelsep=.5em]
    \item \textbf{Analytical/Retrieval queries} ($n=62$): verse/verse-range retrieval and structural filtering (e.g., surah/ayah constraints, juz constraints, revelation class).
    \item \textbf{Counting queries} ($n=74$): aggregation queries (primarily \texttt{COUNT(*)}), optionally combined with filters and ranges.
\end{itemize}
These totals correspond to the denominators used in Table~\ref{tab:nl2sql_benchmark}.

For each sampled query, we used \textbf{GPT-5} to produce a ground-truth SQL query over the schema in Listing~\ref{lst:quran_schema}. This yielded a paired dataset of $(\text{NL query}, \text{SQL}_{gold})$ for automated evaluation.
Direct SQL-string equality is brittle because semantically equivalent SQL can differ syntactically. We therefore evaluate \emph{denotational correctness} by executing both the model prediction ($\text{SQL}_{pred}$) and the ground truth ($\text{SQL}_{gold}$) on the \textbf{same SQLite database} populated with the Quran table:
\begin{itemize}[noitemsep,topsep=0em,leftmargin=1.5em,labelsep=.5em]
    \item For \textbf{scalar} outputs (e.g., counts), correctness requires exact numeric match.
    \item For \textbf{row-valued} outputs, correctness requires equality of returned tuples. Where ordering matters, queries include explicit \texttt{ORDER BY}; otherwise results are compared as unordered sets.
\end{itemize}
A prediction is correct iff the executed result of $\text{SQL}_{pred}$ matches that of $\text{SQL}_{gold}$.
We report accuracy under $N\in\{0,1,5\}$ retrieved in-context examples. For each test query, the system is run with the specified $N$, and correctness is computed using the execution-based protocol above.

\subsubsection{Benchmark Results}
\label{app:nl2sql_results}

Table~\ref{tab:nl2sql_benchmark} reports accuracy (\%) for two baselines (Fanar variants) and the specialized NL2SQL model. The NL2SQL model achieves perfect accuracy on the analytical/retrieval subset across all $N$-shot settings, and shows strong improvements on counting queries as the number of in-context examples increases.

\begin{table}[t]
\centering
\setlength{\tabcolsep}{2pt}
\resizebox{\columnwidth}{!}{
\begin{tabular}{llrrrrrrrr}
\toprule
& & \multicolumn{4}{c}{\textbf{Fanar}} & \multicolumn{4}{c}{\textbf{NL2SQL}} \\
\cmidrule(lr){3-6}\cmidrule(lr){7-10}
\textbf{Task} & $N$
& Correct & Wrong & Total & Acc
& Correct & Wrong & Total & Acc \\
\midrule
Analytical/Retrieval & 0 & 39 & 23 & 62 & 62.90 & 62 & 0 & 62 & 100.00 \\
Analytical/Retrieval & 1 & 50 & 12 & 62 & 80.65 & 62 & 0 & 62 & 100.00 \\
Analytical/Retrieval & 5 & 59 & 3  & 62 & 95.16 & 62 & 0 & 62 & 100.00 \\
\midrule
Counting & 0 & 38 & 36 & 74 & 51.35 & 57 & 17 & 74 & 77.03 \\
Counting & 1 & 59 & 15 & 74 & 79.73 & 66 & 8  & 74 & 89.19 \\
Counting & 5 & 62 & 12 & 74 & 83.78 & 70 & 4  & 74 & 94.59 \\
\bottomrule
\end{tabular}}
\vspace{-0.2cm}
\caption{Execution-based benchmark accuracy (\%) comparing Fanar (with thinking) vs.\ the specialized NL2SQL model on queries sampled from Fanar usage logs. Totals: analytical/retrieval $n=62$, counting $n=74$.}
\label{tab:nl2sql_benchmark}
\end{table}

\subsubsection{NL2SQL Examples}
\label{app:nl2sql_examples}

We include three representative bilingual examples of natural-language Quran queries mapped to executable SQL over the Quran table, spanning verse-range retrieval, specific-verse lookup, and structural statistics.

\paragraph{Example 1: Verse-range retrieval (English).}
\textbf{Query:} What are the first five ayahs of Surah Al-Fatihah?\\
\textbf{SQL:}
\begin{lstlisting}[language=SQL]
SELECT GROUP_CONCAT(text, ' ') AS FullText
FROM (
  SELECT text
  FROM Quran
  WHERE EnglishSurahName = 'Al-Fatiha'
  ORDER BY AyahNumber
  LIMIT 5
);
\end{lstlisting}
\textbf{Result:}
\begin{quote}\AR{\small
بِسْمِ اللَّهِ الرَّحْمَٰنِ الرَّحِيمِ الْحَمْدُ لِلَّهِ رَبِّ الْعَالَمِينَ الرَّحْمَٰنِ الرَّحِيمِ مَالِكِ يَوْمِ الدِّينِ إِيَّاكَ نَعْبُدُ وَإِيَّاكَ نَسْتَعِينُ}
\end{quote}

\paragraph{Example 2: Specific verse retrieval (Arabic).}
\textbf{Query:} \AR{\smallاسترجع الآية 255 من سورة البقرة.} (Surah \textit{Al-Baqarah})\\
\textbf{SQL:}
\begin{lstlisting}[language=SQL]
SELECT text
FROM Quran
WHERE SurahName = Al-Baqarah
  AND AyahNumber = 255;
\end{lstlisting}
\textbf{Result:}
\begin{quote}
\AR{\smallاللَّهُ لَا إِلَٰهَ إِلَّا هُوَ الْحَيُّ الْقَيُّومُ ۚ لَا تَأْخُذُهُ سِنَةٌ وَلَا نَوْمٌ ۚ لَهُ مَا فِي السَّمَاوَاتِ وَمَا فِي الْأَرْضِ ۗ مَنْ ذَا الَّذِي يَشْفَعُ عِنْدَهُ إِلَّا بِإِذْنِهِ ۚ يَعْلَمُ مَا بَيْنَ أَيْدِيهِمْ وَمَا خَلْفَهُمْ ۖ وَلَا يُحِيطُونَ بِشَيْءٍ مِنْ عِلْمِهِ إِلَّا بِمَا شَاءَ ۚ وَسِعَ كُرْسِيُّهُ السَّمَاوَاتِ وَالْأَرْضَ ۖ وَلَا يَئُودُهُ حِفْظُهُمَا ۚ وَهُوَ الْعَلِيُّ الْعَظِيمُ}
\end{quote}

\paragraph{Example 3: Juz-level statistics (Arabic).}

\textbf{Query:} \AR{\smallكم آية في الجزء 30 من القرآن؟}\\
\textbf{SQL:}
\begin{lstlisting}[language=SQL]
SELECT COUNT(*)
FROM Quran
WHERE Juz = 30;
\end{lstlisting}
\textbf{Result:} 564

\subsection{Document Retriever and Embeddings}
\label{app:retriever}






The document retriever performs semantic search over 500K+ documents, including Quran, six major Hadith collections, classical Fiqh texts, contemporary fatwas, Islamic history, and scholarly articles.

The retrieval pipeline embeds each query with Qwen3-Embedding-4B, applies optional query expansion and language detection, and searches a Milvus or Chroma vector index. The index uses HNSW with cosine similarity, top-$k$ retrieval, and a minimum similarity threshold of 0.3. An optional cross-encoder reranker improves precision.

Each retrieved document includes metadata such as source identifier, category, author, language, canonical URL, chunk identifier, and relevance score. The citation module normalizes these records by deduplicating, sorting, extracting metadata, and formatting citations for display.

Qwen3-Embedding-4B produces 4096-dimensional embeddings with an 8192-token context window and supports Arabic, English, and 20+ languages. We use instruction-based embeddings, with task-specific prefixes for queries and no prefix for documents.

For efficiency, the retriever uses an LRU cache for the 1000 most recent embeddings, hash-based deduplication, and persistent vector storage. Batch embedding uses size 32 with parallel execution and progress tracking.

\subsection{Response Assembly and Configuration}
\label{app:response_assembly}



Response assembly uses Server-Sent Events with five event types: \texttt{delta} for tokens, \texttt{citations} for sources and citation flags, \texttt{metadata} for analysis and tool traces, \texttt{status} for progress, and \texttt{done} for completion. Stop-token filtering removes common generation artifacts, including \texttt{<end\_of\_turn>}, \texttt{</s>}, \texttt{<|endoftext|>}, and \texttt{<|im\_end|>}.
Error handling returns bilingual user-facing messages for common failures, including timeouts, retrieval failures, tool failures, and malformed queries.

Configuration uses three-tier hierarchy: database JSON (highest), environment variables, hardcoded defaults. Table~\ref{tab:config_params} presents key parameters.

\begin{table}[h]
\centering
\small
\begin{tabular}{lcc}
\toprule
\textbf{Setting} & \textbf{Default} & \textbf{Range} \\
\midrule
\texttt{greeting.temperature} & 0.2 & 0.0--1.0 \\
\texttt{greeting.max\_tokens} & 256 & 50--1000 \\
\texttt{general.temperature} & 0.1 & 0.0--1.0 \\
\texttt{fiqh.temperature} & 0.1 & 0.0--1.0 \\
\texttt{fiqh.max\_tokens} & 4500 & 2000--12000 \\
\texttt{nl2sql.temperature} & 0.1 & 0.0--0.5 \\
\texttt{max\_sources} & 12 & 5--50 \\
\bottomrule
\end{tabular}
\vspace{-0.2cm}
\caption{Key configuration parameters. Temperature controls randomness, max\_tokens limits length, max\_sources controls retrieval.}
\label{tab:config_params}
\vspace{-0.35cm}
\end{table}

Structured logging outputs JSON records with timestamp, level, component, request ID, truncated query, metadata, latency, and stack traces. Performance metrics track classification, retrieval, inference, per-tool, and end-to-end latency. Error tracking covers fallback events, tool failures, timeouts, retrieval failures, and SQL errors. For Zakat and Inheritance, audit trails record inputs, outputs, assumptions, and warnings for reproducibility.

\section{Evaluation Setups}
We standardized evaluation protocols across all proprietary models. For GPT-4.1 and GPT-5 via OpenAI, we configured inference with medium reasoning effort and standard token limits (1,000 tokens for MCQ, 2,000 for open QA). For Gemini-3 variants (Flash and Pro), we employed medium thinking level settings with temperature 1.0 and comparable token budgets. For Fanar-2-27B and Allam-7B, we used temperature 1.0 with 1,000 token limits. All models received identical task-specific system instructions. MCQ tasks required selection of answer letters without explanation, while open QA tasks requested detailed responses with supporting evidence from Islamic jurisprudence. We implemented few-shot prompting (2 examples) for MCQ evaluation and zero-shot prompting for open QA tasks. No models had access to external tools, retrieval mechanisms, or web search capabilities during evaluation. However, we acknowledge important limitations when comparing proprietary models. Providers do not disclose their exact training data, knowledge cutoff dates, or possible exposure to benchmark datasets. This makes it difficult to determine whether performance differences reflect model capability or memorization of evaluation data. Our evaluation standardizes inference conditions while recognizing these 
limitations.


\end{document}